\documentclass[conference]{IEEEtran}
\IEEEoverridecommandlockouts
\usepackage{cite}
\usepackage{amsmath,amssymb,amsfonts}
\usepackage{algorithmic}
\usepackage{graphicx}
\usepackage{textcomp}
\usepackage{xcolor}
\def\BibTeX{{\rm B\kern-.05em{\sc i\kern-.025em b}\kern-.08em
    T\kern-.1667em\lower.7ex\hbox{E}\kern-.125emX}}
    
\usepackage{amsmath}
\usepackage{dsfont}
\usepackage{tabularx}
\usepackage{color, colortbl}
\definecolor{Gray}{gray}{0.9}
\usepackage{enumitem}
\newcommand{\SubItem}[1]{
    {\setlength\itemindent{15pt} \item[-] #1}
}
\usepackage[flushleft]{threeparttable}

\begin{document}

\title{Reconstructing Missing EHRs Using Time-Aware Within- and Cross-Visit Information for\\ Septic Shock Early Prediction}

%\title{Imputing High Missing EHRs Using Time-Aware Within-Visit and Cross-Visit Information \\ for Septic Shock Early Prediction} 

% \author{Authors \\
% \IEEEauthorblockA{\textit{Organization} \\
% City, Country \\
% email address}
% % \and
% }

\author{\IEEEauthorblockN{1\textsuperscript{st} Ge Gao}
\IEEEauthorblockA{\textit{Department of Computer Science} \\
\textit{North Carolina State University}\\
Raleigh, USA \\
ggao5@ncsu.edu}
\and
\IEEEauthorblockN{2\textsuperscript{nd} Farzaneh Khoshnevisan}
\IEEEauthorblockA{
% \textit{dept. name of organization (of Aff.)} \\
\textit{Intuit Inc.}\\
San Diego, USA \\
farzaneh\_khoshnevisan@intuit.com}
\and
\IEEEauthorblockN{3\textsuperscript{rd} Min Chi}
\IEEEauthorblockA{\textit{Department of Computer Science} \\
\textit{North Carolina State University}\\
Raleigh, USA \\
mchi@ncsu.edu}
}

% \author{\IEEEauthorblockN{Ge Gao\IEEEauthorrefmark{1}, Farzaneh Khoshnevisan\IEEEauthorrefmark{2}, and Min Chi\IEEEauthorrefmark{3}
% }
% \IEEEauthorblockA{
% \IEEEauthorrefmark{1}\IEEEauthorrefmark{3}
% Department of Computer Science,
% North Carolina State University, Raleigh, NC, USA\\
% \IEEEauthorrefmark{2}
% Intuit Inc., San Diego, CA, USA \\
% Email: \IEEEauthorrefmark{1}ggao5@ncsu.edu,
% \IEEEauthorrefmark{2}farzaneh$\_$khoshnevisan@intuit.com,
% and \IEEEauthorrefmark{3}mchi@ncsu.edu}
% }

\maketitle

\begin{abstract}
Real-world Electronic Health Records (EHRs) are often plagued by a high rate of missing data. In our EHRs, for example, the missing rates can be as high as 90\% for some features, with an average missing rate of around 70\% across all features.  We propose a Time-Aware Dual-Cross-Visit missing value imputation method, named \emph{TA-DualCV}, which  spontaneously leverages multivariate dependencies across features and longitudinal dependencies both \textit{within- and cross-visit} to  maximize the information extracted from limited observable records in EHRs. Specifically, TA-DualCV captures the latent structure of missing patterns across measurements of different features and it also considers the time continuity and capture the latent temporal missing patterns based on both time-steps and irregular time-intervals.
TA-DualCV is evaluated using three large real-world EHRs on two types of tasks: an \emph{unsupervised imputation task} by varying mask rates up to 90\% and a \emph{supervised 24-hour early prediction of septic shock} using Long Short-Term Memory (LSTM). Our results show that TA-DualCV performs significantly better than all of the existing state-of-the-art imputation baselines, such as DETROIT and TAME, on both types of tasks. 
\end{abstract}


\section{Introduction}

Sepsis,  defined as life-threatening organ dysfunction in response to infection, is the leading cause of mortality and the most expensive condition associated with in-hospital stay, accounting for more than \$24 billion in annual costs in the United States \cite{liu2014hospital}. In particular, \emph{Septic shock}, the most advanced complication of sepsis due to severe abnormalities of circulation and/or cellular metabolism \cite{bone1992definitions},  reaches a mortality rate as high as 50\% \cite{martin2003epidemiology} and the annual incidence keeps rising \cite{dellinger2008surviving}. It is estimated that as many as 80\% of sepsis deaths could be prevented with early diagnosis and intervention; indeed  prior studies have demonstrated that \emph{early diagnosis} and treatment of septic shock can significantly decrease patients' mortality and shorten their length of stay \cite{kumar2006duration,coba2011resuscitation,henry2015targeted}. Formerly, multiple complex patient health scoring systems have been defined and employed for early diagnosis and early intervention of sepsis, such as SOFA score \cite{vincent1996sofa} and MEDS \cite{shapiro2003mortality}. Despite these efforts, early diagnosis of septic shock is still a challenging problem, because of subtle but fast progression of sepsis/septic shock at early stages with lack of information.  

On the other hand, the rapid growth in volume and diversity of Electronic Health Records (EHRs), makes it possible to apply many machine learning and data mining methods for disease diagnosis. %EHRs presents great potential to guide interventions in public health and facilitates clinical decision-making such as disease management \cite{birkhead2015uses}. 
EHRs are collections of multivariate time series clinical events recorded during patients' visits. \emph{Each visit} consists of a sequence of events pertaining to the changing health status of the patient during the visit. EHRs typically acquire measurements irregularly. For example, vital signs are measured every 8 hours while lab values are measured only every 24 hours. Hence there may not be available readings for lab results when a new event is created for vital signs. Due to this integration of irregular data, EHRs usually contain high missing rates.
 % Generally speaking, there exists a variety of missing mechanisms in EHRs, such as missing completely at random (e.g., equipment failed to collect a patient's data), missing at random (e.g., patients with good vital signs may not have certain lab tests), and missing not at random (e.g., a depressed patient might refuse a depression screening) \cite{little:stat}. As a result, missing rates in EHRs can be high, 
For example, in this work, EHRs collected from general medical system, Christiana Care Health System and Mayo Clinic, have \emph{$73\%$ and $68\%$ missing rates, which are fairly typical of most real-world EHRs}. This missingness in EHRs can severely limit the application of most data mining techniques that require complete data as inputs. 

Our work further differentiates between two types of missing rates: \emph{within-visit missing rate} and \emph{cross-visit missing rate}. Essentially, the former refers to the average value of the missing portion of the data within one visit, while the latter refers to the average value of the missing portion of the data across all visits, and most existing missing rates in literature are cross-visit rates.  Table~\ref{tab:mssingrate} in section~\ref{sec:task1} shows, in EHRs some features have a higher average within-visit missing rate while others have a higher cross-visit missing rate. Our goal here is to  maximize the information extracted from limited observable records in EHRs by spontaneously leverages various dependencies  both \textit{within- and cross-visit}. 

We propose a \textbf{\emph{Time-Aware Dual-Cross-Visit missing value imputation}} approach, named \textbf{\textit{TA-DualCV}}, to impute EHRs with high missing rates. Specifically, it takes advantages of all data points within EHRs by modeling \emph{longitudinal} and \emph{multivariate} dependencies both \emph{within-visit} and \emph{cross-visit}, to resolve the high missing rates in EHRs. \emph{Multivariate dependencies} are captured among measurements across features, and \emph{longitudinal dependencies} consider the time continuity of measurements. Longitudinal dependencies can be captured from two perspectives: \emph{temporal perspective} across time-steps and \emph{time-aware perspective} across time-intervals \emph (i.e., the elapsed time from a starting point till when each event occurs). Each type of dependencies can be modeled either \emph{within-visit} or \emph{cross-visit}. The core part of TA-DualCV is the \emph{dual-cross-visit imputation} (DualCV), with which missingness can be modeled jointly at features and time steps across-visits using chained equations. In particular, it leverages cross-visit information from both feature-perspective and temporal-perspective. Additionally, we design a time-aware imputation (TA) within-visit, which captures longitudinal dependencies across the time intervals. DualCV is combined with TA visit-by-visit to ensure that dependencies are modeled across visits as well as within visits.

Various imputation methods have been proposed previously to handle missingness in EHRs. Based on the purpose of tasks, these prior approaches can be divided into those evaluated on \emph{unsupervised learning tasks} and those designed towards \emph{supervised learning tasks including disease diagnosis}. In the former, different approaches can be categorized into non-Neutral Network (NN)- such as 3D-MICE \cite{luo20183d} and NN-based
approaches such as DeEp impuToR Of mIssing Temporal data (DETROIT) \cite{yan2019detroit}, Time-aware Multi-modal Auto-Encoder (TAME) \cite{yin2020identifying}. Our proposed TA-DualCV is a non-NN based approach using chained equations; as a result, it does not require specific characteristics or assumptions to be followed in EHRs, such as assumptions about the underlying density distributions. For the supervised learning tasks, missing indicator (MI) \cite{little:stat} show great success in disease progression modeling~\cite{ho2014septic, fagerstrom2019lisep, lipton:missing,gao2022gradient}. However, MI cannot be directly used for unsupervised learning tasks.

% Both categories, have achieved great performance in clinical data imputation with moderate missing rate. For unsupervised learning tasks, for example, DETROIT imputes missing values based on neural networks using observed values within a 5-length sliding window and evaluate on EHRs with 7.5\% missing rate. 
% As for missing data imputation approaches for supervised learning tasks, for example, missing indicator (MI) \cite{little:stat} show great success in disease progression modeling~\cite{ho2014septic, fagerstrom2019lisep, lipton:missing}. However, MI cannot be directly used for unsupervised learning tasks.

 Our proposed TA-DualCV can be applied to both unsupervised and supervised learning tasks. Consequently, its effectiveness is assessed against the existing state-of-the-art imputation baselines that can be applied to both types of tasks. For \emph{unsupervised learning tasks}, we assess imputation performance in terms of normalized root-mean-square error (nRMSE), by masking the observations with two commonly used strategies: masking with different rates, and masking one measurement per feature per visit. For \emph{supervised learning tasks}, we utilize the imputed data incorporating Long Short-Term Memory (LSTM) network to predict septic shock 24 hours before its onset. LSTM has been shown to achieve the state-of-the-art results in many real-word applications including disease diagnosis~\cite{khoshnevisan2018recent} through deep hierarchical feature construction. Moreover, it can capture long-range dependencies in time series data in an effective manner~\cite{kim2018temporal}.

Furthermore, the robustness of TA-DualCV is evaluated on EHRs from three different medical systems. Most existing methods are evaluated by using EHRs from single medical system, such as MIMIC-III (Medical Information Mart for Intensive Care) \cite{johnson2016mimic}. As EHR characteristics across different medical systems differ dramatically\cite{khoshnevisan2020adversarial}, it is yet unclear whether existing methods will hold up across different medical systems.  To summarize, our work has at least two main contributions: 
\begin{enumerate}
    \item To handle high missing EHRs, TA-DualCV integrates both cross-visit and within-visit dependencies, by exploiting dependencies among features, time-interval, and time-steps. As far as we know, TA-DualCV is the first \emph{non-NN-based framework} designed for both unsupervised and supervised learning tasks in EHRs. 
    
    \item The generalizability and robustness of TA-DualCV are evaluated on both unsupervised imputation tasks and a supervised task, 24 hours early prediction of septic shock, using EHRs from three different medical systems. The results show that TA-DualCV outperforms state-of-the-art both non-NN- and NN-based baselines in EHRs with high missing rates.
    %Our proposed VACCINE framework does not require assumptions over underlying data characteristics including events numbers across samples, sampling distribution, or substantial observed values. It outperforms imputation baselines on three real-world clinical datasets, containing different types of features and masking rates, for early prediction of septic shock.   
   % \item Our proposed work does not require assumptions over types of missingness (e.g., missing completely at random). As a result, it outperforms imputation baselines on the data with mixed types of missingness.  
\end{enumerate}

\begin{table*}[t]
\caption{Characteristics of Existing Approaches vs. TA-DualCV}

\label{tab:prelim}
\centering
\begin{threeparttable}
% \resizebox{\columnwidth}{!}{
\begin{tabular}{lc|cccc}
\hline
Type & Approach      & Native Missing Rate                       & Tensor Shape                      & Sliding Window               & Prefill                 \\ \hline \hline
Non-NN-Based & MICE (2010)& 17\% & No & No & No \\
& 3D-MICE (2018) & 17\% & No & No & No \\ \hline
NN-Based & BRITS (2018) & 78\% & 	\cellcolor{Gray}Yes & No & \cellcolor{Gray}Yes \\
& DETROIT (2019) & 7\% & 	No & 	\cellcolor{Gray}Yes & \cellcolor{Gray}Yes \\
& GP-VAE (2020) & 78\% & \cellcolor{Gray}Yes &No & No \\
& CATSI (2020) & 7\% & No & No & \cellcolor{Gray}Yes \\
& TAME (2020) & 54\% & \cellcolor{Gray}Yes & No & No \\ \hline
Non-NN-Based & TA-DualCV (Ours)    & 73\% & No & No & No \\ \hline
\end{tabular}
% }
\begin{tablenotes}
\footnotesize
    \item - Native missing rate: it refers to the cross-visit native missing rate of the selected features in the original work. If the approach is evaluated on more than one EHRs, we report the highest one.
    \item - For NN-based approaches, BRITS, CATSI, and TAME are bidirectional NN-based approaches and TAME achieves the best performance. Also, DETROIT is recognized as a strong baseline in~\cite{yin2020identifying}. Thus we use TAME and DETROIT as NN-based baselines in this study. GP-VAE requires tensor-shape input, thus cannot produce results under our EHRs with varied numbers of events across visits.
\end{tablenotes}
\end{threeparttable}
\end{table*}

\section{Related Work}

\subsection{Septic Shock Prediction}

Various machine learning approaches, such as Temporal Belief Memory (TBM) \cite{kim2018temporal} and Time-aware subGroup Basis Approach with Forecasted events (TGBA-F) \cite{yang2018time}, have been applied to missing EHRs to predict septic shock. For example,  TBM \cite{kim2018temporal} captures latent missing patterns based on irregular time intervals of EHRs and imputes missing values within a time window during recurrent neural network (RNN) based classifier training for septic shock prediction. TGBA-F \cite{yang2018time} utilizes matrix decomposition for data imputation by considering forecasting future events, irregular time intervals, and patients subgroups, which is concurrent with RNN classifier training for septic shock prediction. Though both approaches can impute missing values, these need to combine a classifier and cannot be standalone for missing EHRs imputation. As our main focus is constructing a standalone imputing method that can be used for both unsupervised and supervised learning tasks, we don't compare TA-DualCV to these approaches.   

\subsection{Missing EHRs Imputation}

Existing approaches of missing EHRs imputation in general can be divided into \emph{those evaluated on unsupervised learning tasks} and \emph{those designed towards supervised learning tasks}. 

\subsubsection{Approaches evaluated on unsupervised learning tasks} Table~\ref{tab:prelim} summarizes the characteristics of some recently proposed imputation approaches. These approaches can be categorized into \emph{non-Neutral Network (NN)-} and \emph{NN-based approaches}. For example, MICE \cite{buuren2010mice} captures multivariate dependencies using chained equations, and 3D-MICE \cite{luo20183d} combines MICE and Gaussian process to integrate multivariate dependencies cross-visit and time-aware dependencies within-visit. Both MICE and 3D-MICE are non-NN-based and evaluated on MIMIC-III with a moderate missing rate (i.e., 17\%) across their selected 13 features in~\cite{luo20183d}. Benefiting from chained equations, both approaches do not require specific characteristics or assumptions to be followed in EHRs, which motives us to use chained equations for EHRs imputation. 

More recently, NN-based approaches have been proposed and achieved better performance than existing non-NN-based approaches. For example, DETROIT~\cite{yan2019detroit} prefills missing values with means and impute missing values based on neural networks by leveraging multivariate and time-aware dependencies from observed values within a 5-length sliding window, and it is evaluated on a subset of MIMIC-III with a small missing rate (i.e., 7\%) across 13 lab analyte features. GP-VAE~\cite{fortuin2020gpvae} takes tensor-shape data (i.e., a fixed number of events across visits) as model inputs and uses deep variational autoencoders to map the missing data into a latent space without missingness, where it models the low-dimensional dynamics with a Gaussian process. It is evaluated on PhysioNet data~\cite{silva2012predicting} with 78\% missing rate across 35 features. BRITS~\cite{cao2018brits} and CATSI~\cite{yin2020context} prefill missing values and adopt bidirectional recurrent neural networks (RNNs) to model data. BRITS is evaluated on PhysioNet data, and CATSI is evaluated on the same subset of MIMIC-III as DETROIT. BRITS further evaluates their work on in-hospital death classification, however, diseases diagnosis would be more desirable as important supervised learning tasks in healthcare that can guide clinicians construct treatments and intervention. Without prefilling, TAME~\cite{yin2020identifying} utilizes bidirectional RNNs and within-visit multi-modal embedding that takes data including demographics, diagnosis, medication, features, and time-intervals as inputs, and it outputs tensor-shape data without missingness. It is evaluated on MIMIC-III data with 54\% average missing rate across 29 features. 

Despite NN-based approaches' superior performance in imputation, they have at least one of the three major limitations: i) \emph{Tensor-shape input/output}. They require tensor-shape data as input or output, while in clinical practice visits under different health conditions can have a varied number of events. Truncating data into tensor shape can exclude substantial information. ii) \emph{Sliding window}. They impute each missing value using observations within a fixed number of time steps around the missing value, which may not work on highly sparse data. iii) \emph{Prefill}. They need to prefill the missing values to provide fixed-size inputs for NNs, while prefilling can introduce bias which limits model performance. Unlike these approaches, our framework is non-NN-based and does not require tensor-shape input/output, sliding window, or prefilling. BRITS and GP-VAE can handle high missing EHRs. However, they require tensor-shape inputs. As real-world EHRs often contain varied numbers of events across visits, we don't consider BRITS and GP-VAE as baselines in this work. Also, as BRITS, CATSI, and TAME are all bidrectional-RNNs-based among which TAME has achieved the best performance, and DETROIT has been recognized as a strong baseline in~\cite{yan2019detroit}, we use TAME and DETROIT as baselines in this study.              

\subsubsection{Approaches designed towards supervised learning tasks}
Missing indicator (MI) \cite{little:stat} shows great success in disease progression modeling~\cite{ho2014septic,fagerstrom2019lisep,lipton:missing}. However, it only indicates whether a value is observed or missing, thus cannot be directly used for imputation. In this work, we investigate incorporating imputed data with MI for septic shock prediction. Recently, TBM~\cite{kim2018temporal} and TGBA-F~\cite{yang2018time} impute the data concurrent with their classifier training. However, as we have discussed earlier, they cannot be standalone for missing EHRs imputation.

\section{TA-DualCV}

\begin{figure}[t]
    \centering
    \includegraphics[width=9cm]{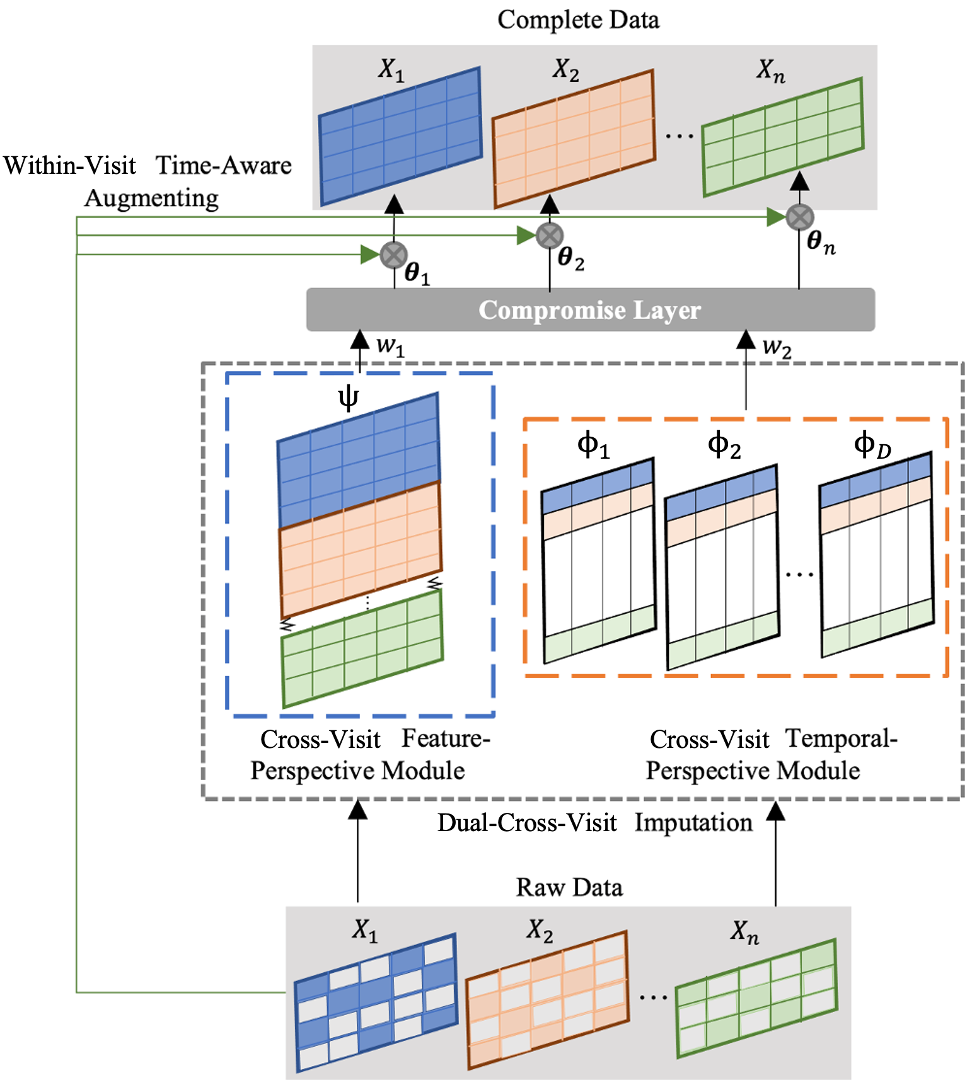}
    \caption{Schematic of TA-DualCV. It consists of two parts: 1) Dual-cross-visit (DualCV) imputation that consists of cross-visits feature-perspective (CFP) and cross-visits temporal-perspective (CTP) modules. We get complete data from a compromise layer which combines two separate results from both modules; 2) Within-visit time-aware imputation (TA). The result from DualCV is augmented visit-by-visit with result from TA.}
    \label{fig:vaccine}
\end{figure}

% \subsection{VACCINE Framework Overview}
Figure~\ref{fig:vaccine} shows the architecture of TA-DualCV.
Specifically, a dual-cross-visit (DualCV) imputation method imputes the missing data using features and temporal dependencies, respectively, then the results from two streams are fused together. The results from DualCV are augmented with GPs which consider the within-visit time-aware dependencies, to constitute the final imputation results. We are releasing the source code of our implementation freely to the research community and it can be accessed at \texttt{\url{https://github.com/fay067/TA-DualCV}}. %Our architecture is designed to handle the following factors and challenges. 

\subsection{Problem Formulation}
\subsubsection{EHRs Imputation}
EHRs can be represented as  $\mathbf{X} = \{\mathbf{X}_1, ..., \mathbf{X}_N\}$, where $N$ is the total number of hospital visits. Each visit $\mathbf{X_i}$ consists of a sequence of events: $\mathbf{X_i} = \{\mathbf{{X}}_{i, 1, \cdot},..., \mathbf{{X}}_{i, T_i, \cdot}\}$ where $\mathbf{{X}}_{i, j, \cdot}\in\mathbb{R}^D$ represents an event recording measurements across $D$ features at time step $j$ in patient $i$'s record. $T_i$ is the number of events (i.e., time-steps) in visit $i$ that can vary across different visits. %Specifically, $\mathbf{{X}}_{i, j, \cdot}\in\mathbb{R}^D$, where $D$ is the number of features (measurements) recorded. 
Each visit $i$ is also associated with a vector $\mathbf{\mathcal{T}}_i = (\tau_{i,1},...,\tau_{i,T_i}) \in\mathbb{R}^{T_i}$ recording the time intervals (e.g., minutes) from a starting point (e.g., patient arrival time) till when each event occurs. Moreover, the events of $\mathbf{X_i}$ are in ascending order of the time intervals. In practice, $\mathbf{X}_i$ may contain unknown \textit{native missing values} due to the irregular data collection process. We apply additional masks to the observed data which introduce \textit{masked missing values} with known groundtruth. Our goal is to devise an imputation model $f$ which generates complete multivariate time series $\mathbf{\hat{X}}$: $\mathbf{\hat{X}} = f(\mathbf{X}, \mathbf{\mathcal{T}})$.

\subsection{DualCV Imputation}
%EHRs from general emergency departments can be highly sparse while existing cross-visit temporal imputation techniques require substantial observed values within neighbors of a missing measurement \cite{xu2020multi}. To address this, DualCV is designed to capture the temporal dependencies underlying missingness, as opposed to existing methods that only consider dependencies across features (e.g., \cite{buuren2010mice}). We adapt chained equations to exploit underlying information from the feature space cross-visit to generate imputations capturing dependencies among features. This approach does not require data to follow specific characteristics, such as the assumptions over underlying density distributions, as the missingness across events and features can be modeled jointly.

DualCV is designed to capture multivariate and temporal dependencies cross-visit, using chained equations. Chained equations are advantageous here because they do not require specific characteristics or assumptions to be followed, such as assumptions about the underlying density distributions. Specifically, we design two chained equation based modules, a CFP module to capture multivariate dependencies cross-visit, as well as a CTP module to capture longitudinal dependencies from temporal perspective cross-visit. We combine the results from the two modules to obtain imputation results, where we call dual-cross-sectional. 
% Instead, it models the missingness across time-steps and features jointly, where we call cross-sectional.    
%The strength of chained equations is it does not require data to follow specific characteristics, such as the assumptions over underlying density distributions, as the missingness across events and features can be modeled jointly.

\subsubsection{Cross-Visits Feature-Perspective Module (CFP)} 
\label{subsec:mice_feature}
To apply CFP, we first transform the data $\mathbf X$ into a single matrix $\mathbf{\Psi} = (\mathbf{X}_{1, \cdot, \cdot},$ $...,\mathbf{X}_{N, \cdot, \cdot})^{\top}$ by concatenating $N$ visits on all features, where we then denote $\mathbf{\Psi}_{j,\cdot}$ as the measurements on $j$-th event and $\mathbf{\Psi}_{\cdot, k}$ represents measurements on $k$-th feature (e.g., white blood cell count), as shown in the Figure \ref{fig:vaccine} (CFP). Each visit can contain different numbers of events and time-intervals between any pair of consecutive events can be different. $\mathbf{\Psi} = (\mathbf{\Psi}_{\cdot, 1},..., \mathbf{\Psi}_{\cdot, D})$ represents the observed part and $\mathbf{\Psi}^{imp} = (\mathbf{\Psi}_{\cdot, 1}^{imp},$ $...,$ $\mathbf{\Psi}_{\cdot, D}^{imp})$ represents the missing part of the visits to be imputed. The predictors of $\mathbf{\Psi}_{\cdot, k}$ is denoted as $\mathbf{\Psi}_{\cdot, -k}=(\mathbf{\Psi}_{\cdot, 1},..., \mathbf{\Psi}_{\cdot, k-1},$ $ \mathbf{\Psi}_{\cdot, k+1},$ $...,$ $\mathbf{\Psi}_{\cdot, D})$ showing the collection of measurements on the $(D-1)$ features except feature $k$. $\mathbf{\Psi}_{\cdot, -k}$ can be incomplete and the correlation between any pair of $\mathbf{\Psi}_{\cdot, k}$ and $\mathbf{\Psi}_{\cdot, -k}$ can be complex (e.g. nonlinear). Therefore, the hypothetically complete data $\mathbf{\hat{\Psi}}=(\mathbf{\hat{\Psi}}_{\cdot, 1},...,\mathbf{{\hat{\Psi}}}_{\cdot, D})$ can be sampled from the $D$-variate multivariate distribution $P(\mathbf{\Psi}_{\cdot, 1},$ $...,\mathbf{\Psi}_{\cdot, D} \mid \gamma)$ where $\gamma$ are parameters \cite{buuren2010mice}. Then the chain equations are designed to get the posterior distribution of $\gamma$ by sampling iteratively from the conditional distribution
\begin{equation} \label{e1.1}
\begin{aligned}
P(\mathbf{\Psi}_{\cdot, 1}\mid \mathbf{\Psi}_{\cdot, -1}, \gamma_1) 	\dots
P(\mathbf{\Psi}_{\cdot, D}\mid \mathbf{\Psi}_{\cdot, -D}, \gamma_D).
\end{aligned}
\end{equation}
% Similar to MICE \cite{buuren2010mice}
Specifically, the $t$-th iteration of chained equations is a Gibbs sampler which draws:
\begin{equation}
\begin{aligned}
& \gamma_1^{imp(t)} \sim P(\gamma_1 \mid \mathbf{\Psi}_{\cdot, 1}, \mathbf{\Psi}_{\cdot, -1}^{t-1}), \\
& \mathbf{\Psi}_{\cdot, 1}^{imp(t)} \sim P(\mathbf{\Psi}_{\cdot, 1} \mid \mathbf{\Psi}_{\cdot, 1}, \mathbf{\Psi}_{\cdot, -1}^{t-1}, \gamma_1^{imp(t)}), \\
& \vdots \\
& \gamma_D^{imp(t)} \sim P(\gamma_D \mid \mathbf{\Psi}_{\cdot, D}, \mathbf{\Psi}_{\cdot, -D}^{t-1}), \\
& \mathbf{\Psi}_{\cdot, D}^{imp(t)} \sim P(\mathbf{\Psi}_{\cdot, D} \mid \mathbf{\Psi}_{\cdot, D}, \mathbf{\Psi}_{\cdot, -D}^{t-1}, \gamma_k^{imp(t)}).
\end{aligned}
\end{equation}
$\mathbf{\hat{\Psi}}_{\cdot, k}^{t} = (\mathbf{\Psi}_{\cdot, k}, \mathbf{\Psi}_{\cdot, k}^{imp(t)})$ is defined as the complete data of $k$-th feature across visits at iteration $t$. Thus, $\mathbf{\hat{\Psi}}^{t} = (\mathbf{\hat{\Psi}}_{\cdot, 1}^{t},...,$ $\mathbf{\hat{\Psi}}_{\cdot, D}^{t})$ is the complete data after $t$-th iteration.

\subsubsection{Cross-Visits Temporal-Perspective Module (CTP)}
\label{subsec:mice_long}
To apply CTP, we then re-formulated the data $\mathbf X$ into $D$ matrices $\mathbf{\Phi}=\{\mathbf{\Phi}_1,...,\mathbf{\Phi}_D\}$ by transposing each visit $\mathbf{X}_i$ and concatenating $N$ visits over the time-steps (e.g., $1,...,T_i$) as shown in Figure ~\ref{fig:vaccine} (CTP). We fix the number of events for each visit to $T^{med}$, which denotes the median value of $\{T_i| i \in [1,N]\}$.  Thus, $\mathbf{\Phi}_j \in \mathbb{R}^{N \times T^{med}}$ denotes the matrix recording measurements on $j$-th feature, where $\mathbf{\Phi}_{j(i,\cdot)}$ represents measurements of $i$-th visit on $j$-th feature and $\mathbf{\Phi}_{j(\cdot,m)}$ represents measurements of the $m$-th time-step on $j$-th feature. If the number of time-steps within a record is less than $T^{med}$, the visit is padded to $T^{med}$ by filling with placeholder values representing missingness (e.g., NaN's). If the number of time-steps is greater than $T^{med}$, the length of the visit is truncated to $T^{med}$ from the front. The reason we use median values is that it is not sensitive to outliers. Also, as the computational complexity of chained-equation based algorithms is directly correlated with the number of variables~\cite{buuren2010mice}, by using median it can balance between efficiency and the number of variables.

Similar to the first step described in Section \ref{subsec:mice_feature}, we apply chained equations on each $\mathbf{\Phi}_j$ separately. Specifically, measurements on the $m$-th time-step of $j$-th feature is denoted by $\mathbf{\Phi}_{j(\cdot,m)}$ while its complementary part is denoted as $\mathbf{\Phi}_{j(\cdot,m)}^{imp}$ which is missing and subjected to imputation. The complete data (i.e., after imputation) on $j$-th feature is denoted as $\mathbf{\hat{\Phi}}_j=(\mathbf{\hat{\Phi}}_{j(\cdot,1)},...,$ $\mathbf{\hat{\Phi}}_{j(\cdot,T^{med})})$ which can be sampled from the $T^{med}$-variate multivariate distribution $P(\mathbf{\Phi}_{j(\cdot,1)},...,\mathbf{\Phi}_{j(\cdot,T^{med})} \mid \beta)$. The conditional distributions used to obtain the posterior distribution of $\mathbf{\Phi}_j$, parametrized by $\beta$, are
\begin{equation}
\begin{aligned}
& P(\mathbf{\Phi}_{j(\cdot, 1)}\mid \mathbf{\Phi}_{j(\cdot, -1)}, \beta_1)
,\dots, \\
& P(\mathbf{\Phi}_{j(\cdot, T^{med})}\mid \mathbf{\Phi}_{j(\cdot, -T^{med})}, \beta_{T^{med}}),
\end{aligned}
\end{equation}
where $\mathbf{\Phi}_{j(\cdot, -m)} = (\mathbf{\Phi}_{j(\cdot, 1)},...,\mathbf{\Phi}_{j(\cdot, m-1)},\mathbf{\Phi}_{j(\cdot, m+1)},...,$ $\mathbf{\Phi}_{j(\cdot, T^{med})})$. Then the Gibbs sampler is used to sample $\beta$ and $\mathbf{\Phi}_j$ iteratively following the chained equations, at the $s$-th iteration, as
\begin{equation}
\small
\begin{aligned}
& \beta_1^{imp(s)} \sim P(\beta_1 \mid \mathbf{\Phi}_{j(\cdot, 1)}, \mathbf{\Phi}_{j(\cdot, -1)}^{s-1}), \\
& \mathbf{\Phi}_{j(\cdot, 1)}^{imp(s)} \sim P(\mathbf{\Phi}_{j(\cdot, 1)} \mid \mathbf{\Phi}_{j(\cdot, 1)}, \mathbf{\Phi}_{j(\cdot, -1)}^{s-1}, \beta_1^{imp(s)}), \\
& \vdots \\
& \beta_{T^{med}}^{imp(s)} \sim P(\beta_{T^{med}} \mid \mathbf{\Phi}_{j(\cdot, T^{med})}, \mathbf{\Phi}_{j(\cdot, -T^{med})}^{s-1}), \\
& \mathbf{\Phi}_{j(\cdot, T^{med})}^{imp(s)} \sim P(\mathbf{\Phi}_{j(\cdot, T^{med})} \mid \mathbf{\Phi}_{j(\cdot, T^{med})}, \mathbf{\Phi}_{j(\cdot, -T^{med})}^{s-1}, \beta_{T^{med}}^{imp(s)}).
\end{aligned}
\end{equation}
$\mathbf{\hat{\Phi}}_{j(\cdot, m)}^{s} = (\mathbf{\Phi}_{j(\cdot, m)}, \mathbf{\Phi}_{j(\cdot, m)}^{imp(s)})$ is the complete data on $j$-th feature $m$-th time-step at iteration $s$. Then the complete (i.e., imputed) data on $j$-th feature, after the $s$-th iteration, is obtained as $\mathbf{\hat{\Phi}}_j^{s} = (\mathbf{\hat{\Phi}}_{j(\cdot, 1)}^{s},...,\mathbf{\hat{\Phi}}_{j(\cdot, T^{med})}^{s})$.

\subsubsection{Compromising Layer}
We now have two complete results, $\mathbf{\hat{\Psi}}$ and $\mathbf{\hat{\Phi}}$, obtained using CFP and CTP which considers the dependencies captured from the feature and temporal space respectively. Then $\mathbf{\hat{\Psi}}$ and $\mathbf{\hat{\Phi}}$ are fused together to constitute the complete data $\mathbf{\hat{X}}^{M}$. Specifically, the value on $i$-th visit, $j$-th time-step, and $k$-th feature of $\mathbf{\hat{X}}^{M}$, $\hat{X}_{i,j,k}^{M}$, is obtained as following
\begin{equation}
\begin{aligned}
\hat{X}_{i,j,k}^{M} = [w_1\hat{\Psi}_{\sum_{x=1}^{i-1}T_x, k}^t + w_2\hat{\Phi}_{k(i,j)}^s]\mathds{1}\{T_i \le T^{med}\} \\
+ \hat{\Psi}_{\sum_{x=1}^{i-1}T_x, k}^t\mathds{1}\{T_i > T^{med}\},
\end{aligned}
\end{equation}
where $\mathds{1}$ is indicator function, $w_1$ and $w_2$ are hyperparameters \textit{s.t.} $w_1 + w_2 = 1$.

\subsection{Within-Visit Time-Aware Augmenting Mechanism}
%Another challenge in time series data imputation is the trade-off between cross-visit and within-visit dependencies. For example, a measurement from one visit can be modeled using neighboring observations within the visit (e.g., across different time-steps), or using the same feature obtained at a similar time-step across other visits in the sample population. To balance this, we design time-aware imputation within each visit by GP, which shows its success in capturing temporal dependencies of clinical data in existing works \cite{luo20183d}. It augments the imputations output from DualCV accordingly.
Besides DualCV, which captures the feature and temporal dependencies \textit{across} visits, we employ GP to perform time-aware imputations on each visit to capture patient-specific correlations \textit{within} each visit. The difference between CTP and within-visit time-aware imputation is that CTP focuses on ``outer" dependencies among time-steps of events across samples within population, while GP captures the ``inner" dependencies among time-intervals of events within each visit. 

Specifically, we apply single-task GP imputation over each feature on visit $\mathbf{X}_i$ individually. % For $\mathbf{X}_i$, a univariate time series $\mathbf{\mathcal{T}}_{i}=(\tau_{i,1},...,\tau_{i, T_i})$ is a $T_i$-dimensional vector generated with each visit $i$. 
% The simulator output $y(\mathbf{\mathcal{T}}_{i})$ is denoted as: 
% \begin{equation}\label{GP:1}
%     y(\mathbf{\mathcal{T}}_{i})=\mu + z(\mathbf{\mathcal{T}}_{i})
% \end{equation}
% where $\mu$ is overall mean and $z(\mathbf{\mathcal{T}}_{i})$ is a GP that we assume to reach a local optima on its covariance structure:
% \begin{equation}\label{GP:2}
%     Cov(z(\mathbf{\mathcal{T}}_{i}), z(\mathbf{\mathcal{T}}_{j}))=\sigma_{z}^2 R_{ij}
% \end{equation}
% In general, $y(\mathbf{\mathcal{T}}) = (y(\mathbf{\mathcal{T}}_{1}),...,y(\mathbf{\mathcal{T}}_D))^{\top}$ has a multivariate normal distribution. 
% We specifically follow the Gaussian correlation function defined by Ranjan et al.~\cite{ranjan2011computationally}:
% \begin{equation}
%     R_{ij} = \prod_{k=1}^{t_i} exp\{-\alpha_{k} \mid \tau_{i,k}-\tau_{j,k} \mid ^2\}  \mbox{ for all }  i,j
% \label{eq:R-matrix}
% \end{equation}
% where $\alpha=(\alpha_1,...,\alpha_{t_i})$ is a vector of hyperparameters.
The observed data on the $l$-th feature of visit $\mathbf{X}_i$ is denoted as $\mathbf{X}_{i,\cdot,l}$, with $\mathbf{X}_{i,\cdot,l}^{imp}$ representing its complementary missing part which is subject to be imputed. 
For $l$-th feature, we split the time-interval vector $\mathbf{\mathcal{T}}_i^{(l)}$ into $\mathbf{\mathcal{T}}_i^{obs(l)}$ and $\mathbf{\mathcal{T}}_i^{imp(l)}$, where $\mathbf{\mathcal{T}}_i^{obs(l)}$ and $\mathbf{\mathcal{T}}_i^{imp(l)}$ denote the time-intervals corresponding to $\mathbf{X}_{i,\cdot,l}$ and $\mathbf{X}_{i,\cdot,l}^{imp}$ respectively.
The target output $\mathbf{X}_{i,\cdot,l}^{imp}$ is modeled as:
\begin{equation}\label{GP:1}
    \mathbf{X}_{i,\cdot,l}^{imp}=\mu + z(\mathbf{\mathcal{T}}_i^{imp(l)})
\end{equation}
where $\mu$ is overall mean and $z(\mathbf{\mathcal{T}}_i^{imp(l)})$ is a GP that we assume to reach a local optima on its covariance structure:
\begin{equation}\label{GP:2}
    Cov(z(\mathbf{\mathcal{T}}_i^{imp(l)}), z(\mathbf{\mathcal{T}}_i^{imp(m)}))=\sigma_{z}^2 R_{lm}
\end{equation}
In general, $\mathbf{X}_{i,\cdot,l}^{imp} = (\mathbf{X}_{i,\cdot,1}^{imp},...,\mathbf{X}_{i,\cdot,D}^{imp})^{\top}$ has a multivariate normal distribution. 
We specifically follow the Gaussian correlation function defined by Ranjan et al.~\cite{ranjan2011computationally}:
\begin{equation}
    R_{lm} = \prod_{k=1}^{t_i} exp\{-\alpha_{k} \mid \mathbf{\mathcal{T}}_{i,k}^{imp(l)}-\mathbf{\mathcal{T}}_{i,k}^{imp(m)} \mid ^2\}  \mbox{ for all }  l,m
\label{eq:R-matrix}
\end{equation}
where $\alpha=(\alpha_1,...,\alpha_{t_i})$ is a vector of hyperparameters.

According to the Equations \ref{GP:1}, \ref{GP:2}, and \ref{eq:R-matrix}, we can use $(\mathbf{\mathcal{T}}_i^{obs(l)}, \mathbf{X}_{i,\cdot,l}^{obs})$ to fit GP and estimate the $\alpha$ following the negative profile log-likelihood \cite{macdonald2015gpfit} 
\begin{equation}
\small
\begin{aligned}
-2 & log(L_\alpha)  \propto \\
& log(\mid R\mid) + nlog [(\mathbf{X}_{i,\cdot,l}^{obs} - \mathbf{1_n}\hat{\mu}(\alpha))^\top R^{-1}(\mathbf{X}_{i,\cdot,l}^{obs} - \mathbf{1_n}\hat{\mu}(\alpha))],
\end{aligned}
\end{equation}
where $\mid$$R$$\mid$ denotes the determinant of the Gaussian correlation function R defined in~\eqref{eq:R-matrix}. Then we can get the predicted $\mathbf{X}_{i,\cdot,l}^{imp}$ by evaluating the GP model with input $\mathbf{\mathcal{T}}_i^{imp(l)}$. Finally, we can obtain the complete (i.e., imputed) data of visit $i$ as $\mathbf{\hat{X}}_i^G = (\mathbf{{X}}_i, \mathbf{X}_i^{imp})$ which constitutes the imputed dataset $\mathbf{\hat{X}}^G=(\mathbf{\hat{X}}_1^G,...,\mathbf{\hat{X}}_N^G)$.

Finally, we augment the results from DualCV with the outputs from GP through weighted averaging $\mathbf{\hat{X}}_i = \mathbf{\theta}_{i,1}\cdot\mathbf{\hat{X}}_i^M +  \mathbf{\theta}_{i,2}\cdot\mathbf{\hat{X}}_i^G$, where $\mathbf{\theta}_{i,1}=\frac{\sigma_{G_i}}{\sigma_{M_i} + \sigma_{G_i}}$,  $\mathbf{\theta}_{i,2}=\frac{\sigma_{M_i}}{\sigma_{M_i} + \sigma_{G_i}}$, $\sigma_{M_i}$ and $\sigma_{G_i}$ are the standard deviations of the imputations output from DualCV and within-visit time-aware imputations given visit $i$, respectively. Note the weight is designed to penalize the results (either from DualCV or GP) that are associated with the larger standard deviations.

\section{Imputation with High Missing Rates}
\label{sec:task1}

\begin{table}[]
\caption{Missing rates of overall 22 features in CCHS, Mayo, and MIMIC-III. }
\label{tab:mssingrate}
\begin{threeparttable}
\begin{tabularx}{\columnwidth}{cccc|ccc}
\hline
\multicolumn{1}{c|}{Feature} & \multicolumn{3}{c|}{Cross-Visit} & \multicolumn{3}{c}{Within-Visit} \\
\multicolumn{1}{c|}{}                    & MIMIC     & CCHS            & Mayo           &MIMIC           & CCHS     & Mayo      \\ \hline
\multicolumn{1}{c|}{CI}             & 26\%                &                 &         & 26\%        &          &    \\
\multicolumn{1}{c|}{K}                     & 21\%               &                 &                 & 20\%     &                 &          \\
\multicolumn{1}{c|}{HCO3}                   & 27\%                   &                 &          & 27\%         &          &     \\ 
\multicolumn{1}{c|}{Na}            & 24\%                       &                 &     & 24\%          &          &    \\
\multicolumn{1}{c|}{Hct}                 & 18\%                      &                 &                 & 19\%          &                 &          \\
\multicolumn{1}{c|}{Hb}             & 30\%                     &                 &              & 30\%
&          &   \\
\multicolumn{1}{c|}{MCV}            & 31\%              &                &     & 31\%          &          &   \\
\multicolumn{1}{c|}{PLT}           & 28\%                      &            &     & 28\%           &     &    \\
\multicolumn{1}{c|}{WBC}            & 30\%                      &            &     & 30\%          &     &    \\
\multicolumn{1}{c|}{RDW}                  & 31\%                      &               &         & 30\%          &          &    \\ 
\multicolumn{1}{c|}{BUN}           & 26\%                      &            &     & 25\%          &     &   \\
\multicolumn{1}{c|}{Cr}           & 25\%                       &            &     & 25\%          &     &    \\
\multicolumn{1}{c|}{GLC}            & 30\%                      &                 &     & 30\%          &         &   \\ \hline
\multicolumn{1}{c|}{Temp}           & 97\%                     & 81\%           & 91\%           & 95\%          & 82\%           & 88\%          \\
\multicolumn{1}{c|}{RespRate}           & 45\%                 & 63\%          & 49\%           & 61\%          & 65\%           & 54\%          \\
\multicolumn{1}{c|}{HeartRate}           & 44\%                & 61\%           & 54\%           & 61\%           & 64\%           & 55\%         \\

\multicolumn{1}{c|}{FiO2}           & 84\%                     & 85\%           & 94\%           & 96\%          & 84\%           & 91\%          \\

\multicolumn{1}{c|}{PulseOx}           & 45\%                 & 64\%           & 34\%            & 62\%        & 68\%           & 40\%          \\
\multicolumn{1}{c|}{OFlow}           & 94\%                    & 78\%           & 97\%            & 94\%         & 87\%           & 92\%          \\
\multicolumn{1}{c|}{DBP}           & 33\%                      & 72\%           & 63\%           & 25\%          & 82\%           & 64\%          \\ 
\multicolumn{1}{c|}{SBP}           & 33\%                      & 72\%           & 63\%           & 25\%          & 72\%           & 64\%          \\
\multicolumn{1}{c|}{MAP}           & 33\%                      & 77\%           & 63\%            & 25\%         & 77\%           & 68\%          \\\hline
\multicolumn{1}{c|}{Median}           & 31\%      & 72\%           & 63\%          & 29\%          & 77\%           & 64\%           \\ 
\multicolumn{1}{c|}{Mean}           & 39\%      & 73\%           & 68\%           & 40\%          & 76\%           & 68\%          \\ \hline
\end{tabularx}
\begin{tablenotes}
      \scriptsize
      \item - The median/mean refers to the median/average missing rate of the \emph{selected features}, which can be different from the median/average missing rate across all features.
    \end{tablenotes}
\end{threeparttable}
\end{table}

\begin{table*}[tbp]
% \scriptsize
% \small
\centering
\caption{Imputation results (nRMSE) with masking rates 60\%-90\% on 13 features of MIMIC-III.}
\label{tab:nrmse}
\resizebox{\textwidth}{!}{%
\begin{threeparttable}
\begin{tabular}{lcccccccccccccc}
\hline                                                                                                                                                                                                                                                                                                                                                                                \\ \hline
\multicolumn{1}{l|}{Model} & \multicolumn{1}{c}{CI} & \multicolumn{1}{c}{K} & \multicolumn{1}{c}{HCO3} & \multicolumn{1}{c}{Na} & \multicolumn{1}{c}{Hct} & \multicolumn{1}{c}{Hb} & \multicolumn{1}{c}{MCV} & \multicolumn{1}{c}{PLT} & \multicolumn{1}{c}{WBC} & \multicolumn{1}{c}{RDW} & \multicolumn{1}{c}{BUN} & \multicolumn{1}{c}{Cr} & \multicolumn{1}{c}{GLC} & \multicolumn{1}{c}{Average} \\ \hline

\multicolumn{14}{l}{Mask rate: 90\%}                                                                                                                                                               \\ \hline
\multicolumn{1}{l|}{MeanFill} & 1.296 & 1.208 & 1.143 & 1.120 & 4.227 & 2.729 & 2.337 & 8.754 & 5.065 & 4.721 & 3.368 & 3.398 & 4.488 & 3.373\\
\multicolumn{1}{l|}{ECF}                          & \textbf{0.270}                        & \textbf{0.256}                         & \textbf{0.272}                      & \textbf{0.267}                      & \textbf{0.275}                          & \textbf{0.278}                          & \textbf{0.295}                   & \textbf{0.294}                              & \textbf{0.279}                   & \textbf{0.304}                   & \textbf{0.283}                   & \textbf{0.280}                          & \textbf{0.260}                   & \textbf{0.278}    \\ \cline{2-15}
\multicolumn{1}{l|}{MICE}       & 0.283                        & 0.269                         & 0.287                      & 0.280                      & 0.282                          & 0.284                         & 0.310                   & 0.310                              & 0.295                   & 0.321                   & 0.304                   & 0.318                         & 0.271           & 0.293            \\ 
\multicolumn{1}{l|}{3D-MICE}    & \textbf{0.259}                        & \textbf{0.253}                         & \textbf{0.262}                      & \textbf{0.259}                      & \textbf{0.262}                 & \textbf{0.263}                 & \textbf{0.270}                  & \textbf{0.274}                              & \textbf{0.270}                   & \textbf{0.265}                   & \textbf{0.267}                   & \textbf{0.261}                          & \textbf{0.256}                  & \textbf{0.263}     \\  \cline{2-15}
\multicolumn{1}{l|}{DETROIT} & 4.457 & \textbf{0.289} & \textbf{0.311} & 2.855 & 0.421 & \textbf{0.371} & 0.981 & 0.578 & 2.121 & \textbf{0.303} & 1.720 & 2.014 & 1.585 & 1.385\\
\multicolumn{1}{l|}{TAME} & \textbf{0.381} & 0.388 & 0.318 & \textbf{0.323} & \textbf{0.387} & 0.380 & \textbf{0.734} & \textbf{0.436} & \textbf{0.465} & 0.441 & \textbf{0.482} & \textbf{1.290} & \textbf{0.365} & \textbf{0.491} \\ \hline
\multicolumn{1}{l|}{TA-DualCV$^{-C}$}                              & 0.284                        & 0.276                         & 0.285                      & 0.284                      & 0.288                          & 0.290                          & 0.296                   & 0.300                              & 0.296                   & 0.289                   & 0.292                   & 0.284                          & 0.279         & 0.288              \\
\multicolumn{1}{l|}{TA-DualCV$^{-I}$} & 0.254 & 0.264 & 0.255 & 0.257 & 0.258 & 0.261 & 0.266 & 0.267 & 0.277 & 0.257 & 0.257 & 0.260 & 0.267 & 0.261 \\
\multicolumn{1}{l|}{TA-DualCV}    & \textbf{0.221}**               & \textbf{0.225}**                & \textbf{0.224}**             & \textbf{0.222}**             & \textbf{0.229}**                          & \textbf{0.233}**                          & \textbf{0.232}**          & \textbf{0.236}**                     & \textbf{0.232}**          & \textbf{0.237}**          & \textbf{0.229}**          & \textbf{0.223}**                 & \textbf{0.218}**        & \textbf{0.228}**       \\ \hline

\multicolumn{14}{l}{Mask rate: 80\%}                                                                                                            \\ \hline
\multicolumn{1}{l|}{MeanFill} & 1.201 & 1.105 & 1.065 & 1.072 & 4.325 & 2.486 & 2.174 & 7.354 & 4.189 & 4.394 & 3.032 & 3.211 & 3.878 & 3.037\\
\multicolumn{1}{l|}{ECF}     & \textbf{0.267}                        & \textbf{0.254}                        & \textbf{0.268}                      & \textbf{0.265}                      & \textbf{0.273}                          & \textbf{0.276}                          & \textbf{0.292}                   & \textbf{0.289}                              & \textbf{0.276}                   & \textbf{0.299}                   & \textbf{0.279}                   & \textbf{0.276}                          & \textbf{0.258}        & \textbf{0.275}               \\ \cline{2-15} 
\multicolumn{1}{l|}{MICE}                            & 0.287                        & 0.270                         & 0.290                      & 0.283                      & 0.286                          & 0.289                          & 0.312                   & 0.313                              & 0.297                   & 0.322                   & 0.307                   & 0.324                          & 0.273             & 0.296          \\ 
\multicolumn{1}{l|}{3D-MICE}                         & \textbf{0.257}                        & \textbf{0.250}                         & \textbf{0.259}                      & \textbf{0.256}                      & \textbf{0.259}                          & \textbf{0.261}                          & \textbf{0.267}                   & \textbf{0.273}                              & \textbf{0.267}                   & \textbf{0.264}                   & \textbf{0.264}                   & \textbf{0.258}                          & \textbf{0.253}       & \textbf{0.260}                \\ \cline{2-15}
\multicolumn{1}{l|}{DETROIT} & \textbf{0.272} & \textbf{0.257} & \textbf{0.256} & \textbf{0.257} & \textbf{0.256} & \textbf{0.251} & \textbf{0.407} & \textbf{0.287} & \textbf{0.321} & \textbf{0.270} & \textbf{0.314} & 1.544 & \textbf{0.243} & \textbf{0.380}\\
\multicolumn{1}{l|}{TAME} & 0.317 & 0.294 & 0.314 & 0.317 & 0.389 & 0.402 & 0.513 & 0.527 & 0.602 & 0.624 & 0.343 & \textbf{0.484} & 0.389 & 0.424\\ \hline
\multicolumn{1}{l|}{TA-DualCV$^{-C}$}         & 0.277                        & 0.272                         & 0.280                     & 0.279                     & 0.283                          & 0.284                         & 0.290                   & 0.292                              & 0.290                   & 0.280                   & 0.286                  & 0.279                          & 0.274             & 0.282          \\
\multicolumn{1}{l|}{TA-DualCV$^{-I}$} & 0.246 & 0.251 & 0.250 & 0.247 & 0.242 & 0.242 & 0.268 & 0.262 & 0.262 & 0.271 & 0.256 & 0.259 & 0.257 & 0.255 \\
\multicolumn{1}{l|}{TA-DualCV}                         &\textbf{0.222}**               & \textbf{0.219}**                & \textbf{0.224}**             & \textbf{0.222}**             & \textbf{0.228}**                 & \textbf{0.231}**                 & \textbf{0.235}**          & \textbf{0.236}**                     & \textbf{0.233}**          & \textbf{0.234}**          & \textbf{0.225}**          & \textbf{0.226}**                 & \textbf{0.221}**    &  \textbf{0.227}**         \\ \hline

\multicolumn{14}{l}{Mask rate: 70\%}                                                                                                                                                                                                                                                                                                                                                                                                   \\ \hline
\multicolumn{1}{l|}{MeanFill} & 0.847 & 0.787 & 0.789 & 0.764 & 2.611 & 1.752 & 1.726 & 4.394 & 3.225 & 3.247 & 2.170 & 2.674 & 2.159 & 2.088\\
\multicolumn{1}{l|}{ECF}     & \textbf{0.255}                        & \textbf{0.249}                         & \textbf{0.258}                      & \textbf{0.255}                      & \textbf{0.264}                          & \textbf{0.266}                          & \textbf{0.279}                   & \textbf{0.277}                              & \textbf{0.264}                   & \textbf{0.285}                   & \textbf{0.265}                   & \textbf{0.264}                          & \textbf{0.254}           & \textbf{0.264}            \\ \cline{2-15} 
\multicolumn{1}{l|}{MICE}       & 0.271                        & 0.264                         & 0.279                      & 0.269                      & 0.264                          & 0.265                          & 0.304                   & 0.302                              & 0.287                   & 0.314                   & 0.293                   & 0.310                          & 0.268            & 0.284           \\ 

\multicolumn{1}{l|}{3D-MICE}    & \textbf{0.243}                        & \textbf{0.242}                         & \textbf{0.248}                      & \textbf{0.245}                      & \textbf{0.244}                 & \textbf{0.243}                 & \textbf{0.251}                  & \textbf{0.256}                              & \textbf{0.254}                   & \textbf{0.246}                   & \textbf{0.250}                   & \textbf{0.245}                          & \textbf{0.246}   & \textbf{0.247}                    \\ \cline{2-15}
\multicolumn{1}{l|}{DETROIT} & \textbf{0.239} & \textbf{0.244} & 0.301 & 0.266 & \textbf{0.227} & \textbf{0.220}** & \textbf{0.258} & 0.361 & 0.449 & \textbf{0.238}** & 0.790 & 0.849 & \textbf{0.263} & 0.362 \\
\multicolumn{1}{l|}{TAME} & 0.274 & 0.320 & \textbf{0.281} & \textbf{0.265} & 0.237 & 0.235 & 0.497 & \textbf{0.326} & \textbf{0.370} & 0.410 & \textbf{0.330} & \textbf{0.509} & 0.444 & \textbf{0.346}\\ \hline
\multicolumn{1}{l|}{TA-DualCV$^{-C}$}         & 0.255                        & 0.249                         & 0.258                      & 0.255                      & 0.264                          & 0.266                         & 0.279                   & 0.277                              & 0.264                   & 0.285                   & 0.265                  & 0.264                          & 0.254       & 0.264                \\
\multicolumn{1}{l|}{TA-DualCV$^{-I}$} & 0.231 & 0.235 & 0.240 & 0.234 & 0.223 & 0.225 & 0.247 & 0.248 & 0.256 & 0.246 & 0.238 & 0.242 & 0.247 & 0.239  \\
 \multicolumn{1}{l|}{TA-DualCV}    &   \textbf{0.220}** & \textbf{0.221}** & \textbf{0.221}** & \textbf{0.229}** & \textbf{0.221}** & \textbf{0.222} & \textbf{0.235}** & \textbf{0.231}** & \textbf{0.229}** & \textbf{0.239} & \textbf{0.225}** & \textbf{0.224}** & \textbf{0.214}** & \textbf{0.225}**        \\ \hline

\multicolumn{14}{l}{Mask rate: 60\%}                                                                                                                                                                                                                                                                                                                                                                                                   \\ \hline
\multicolumn{1}{l|}{MeanFill}  & 0.657 & 0.589 & 0.624 & 0.591 & 1.801 & 1.279 & 1.544 & 2.958 & 1.782 & 2.538 & 1.685 & 2.370 & 1.601 & 1.540 \\
\multicolumn{1}{l|}{ECF}     & \textbf{0.247}                        & \textbf{0.246}                         & \textbf{0.250}                      & \textbf{0.248}                      & \textbf{0.258}                          & \textbf{0.260}                          & \textbf{0.272}                   & \textbf{0.266}                              & \textbf{0.257}                   & \textbf{0.275}                   & \textbf{0.253}                   & \textbf{0.255}                          & \textbf{0.252}  & \textbf{0.257}                      \\ \cline{2-15}
\multicolumn{1}{l|}{MICE}       & 0.260                        & 0.261                         & 0.271                      & 0.259                      & 0.247                          & 0.247                          & 0.301                   & 0.298                              & 0.283                   & 0.310                   & 0.285                   & 0.296                          & 0.266  & 0.276                     \\ 

\multicolumn{1}{l|}{3D-MICE}    & \textbf{0.233}                        & \textbf{0.234}                         & \textbf{0.239}                      & \textbf{0.235}                      & \textbf{0.229}                 & \textbf{0.228}                 & \textbf{0.240}                  & \textbf{0.244}                              & \textbf{0.245}                   & \textbf{0.232}                   & \textbf{0.238}                   & \textbf{0.236}                          & \textbf{0.238}   & \textbf{0.236}                    \\ \hline
\multicolumn{1}{l|}{DETROIT} & \textbf{0.154}** & \textbf{0.241} & \textbf{0.168}** & \textbf{0.154}** & \textbf{0.149}** & \textbf{0.196}** & \textbf{0.164}** & 0.481 & 0.458 & \textbf{0.154}** & \textbf{0.216}** & \textbf{0.411} & \textbf{0.170}** & \textbf{0.240} \\
\multicolumn{1}{l|}{TAME} & 0.227 & 0.271 & 0.257 & 0.245 & 0.211 & 0.212 & 0.355 & \textbf{0.321} & \textbf{0.320} & 0.352 & 0.270 & 0.423 & 0.321 & 0.291 \\ \cline{2-15}
\multicolumn{1}{l|}{TA-DualCV$^{-C}$} & 0.271 & 0.275 & 0.271 & 0.274 & 0.281 & 0.283 & 0.284 & 0.281 & 0.285 & 0.269 & 0.271 & 0.275 & 0.282 & 0.277 \\  
\multicolumn{1}{l|}{TA-DualCV$^{-I}$} & \textbf{0.218} & 0.237 & 0.230 & \textbf{0.223} & \textbf{0.205} & \textbf{0.207} & 0.260 & 0.245 & 0.247 & 0.256 & 0.234 & 0.240 & 0.249 & 0.235 \\ 
\multicolumn{1}{l|}{TA-DualCV} &  0.219 & \textbf{0.223}** & \textbf{0.220} & 0.224 & 0.218 & 0.218 & \textbf{0.229} & \textbf{0.230}** & \textbf{0.223} & \textbf{0.222} & \textbf{0.225} & \textbf{0.223}** & \textbf{0.209} & \textbf{0.222}** \\ \hline
\end{tabular}
\begin{tablenotes}
      \small
      \item - For each block, the best approach is in \textbf{bold}; The best approach across ALL is labeled with **.
    \end{tablenotes}
\end{threeparttable}
}
\end{table*}

% \begin{table*}[!htbp]
% \small
% \centering
% \caption{Imputation results (in nRMSE) on 9 sepsis-related features of CCHS, Mayo, and MIMIC-III.}
% \label{tab:nrmse-sepsis}
% \begin{tabular}{c|cc|cccc|c}
% \hline
% Dataset   & MeanFill & ECF   & MICE  & 3D-MICE & DETROIT & TAME  & TA-DualCV         \\ \hline
% CCHS      & \textbf{2.285}    & 2.785 & 1.043 & \textbf{0.984}   & N/A     & 0.995 & \textbf{0.285}* \\
% Mayo      & 7.109    & \textbf{6.942} & 3.617 & \textbf{3.592}   & N/A     & 4.806 & \textbf{0.305}* \\
% MIMIC-III & 1.455    & \textbf{1.231} & 0.829 & \textbf{0.823}   & N/A     & 0.914 & \textbf{0.289}* \\ \hline
% \end{tabular}
% \end{table*}

\begin{table*}[!htbp]
% \scriptsize
\small
% \fontsize{3}{6}\selectfont
\centering
\caption{Imputation results (nRMSE) with classic masking rates on 9 sepsis-related features of CCHS, Mayo, and MIMIC-III.}
\label{tab:nrmse-sepsis}
% \resizebox{\textwidth}{!}{%
\begin{threeparttable}
\begin{tabular}{lcccccccccc}
\hline
\multicolumn{1}{l|}{Model} & \multicolumn{1}{l}{Temp} & \multicolumn{1}{l}{RespRate} & \multicolumn{1}{l}{HeartRate} & \multicolumn{1}{l}{FiO2} & \multicolumn{1}{l}{PulseOx} & \multicolumn{1}{l}{OFlow} & \multicolumn{1}{l}{DBP} & \multicolumn{1}{l}{SBP} & \multicolumn{1}{l}{MAP} & \multicolumn{1}{l}{Average} \\ \hline

\multicolumn{11}{l}{\textbf{CCHS}}  \\ \hline
\multicolumn{1}{l|}{MeanFill} &  1.929 & 0.624 & 1.704 & \textbf{2.267} & \textbf{0.655} & \textbf{12.927} & \textbf{1.092} & \textbf{1.404} & \textbf{0.867} & 2.825  \\  
\multicolumn{1}{l|}{ECF}     & \textbf{1.872} & \textbf{0.564} & \textbf{1.498} & \textbf{2.267} & \textbf{0.655} & \textbf{12.927} & \textbf{1.092} & \textbf{1.404} & \textbf{0.867} & \textbf{2.785}\\ \cline{2-11} 
\multicolumn{1}{l|}{MICE}    & 1.224 & 0.644 & 0.744 & 0.391 & 0.572 & 0.557 & 2.564 & 1.644 & 1.268 & 1.043\\
\multicolumn{1}{l|}{3D-MICE} & 1.219 & 0.637 & \textbf{0.714} & \textbf{0.311} & \textbf{0.549} & \textbf{0.473} & 2.485 & 1.483 & \textbf{1.179} & \textbf{0.984} \\
\multicolumn{1}{l|}{TAME}    & \textbf{0.808} & \textbf{0.631} & 0.986 & 0.980 & 0.880 & 1.484 & \textbf{0.790} & \textbf{1.401} & 1.801 & 1.085\\ \hline

\multicolumn{1}{l|}{TA-DualCV$^{-C}$}     & 0.355 & \textbf{0.218}** & 0.487 & 0.343 & 0.374 & \textbf{0.236}** & 0.303 & 0.281 & 0.222 &  0.313 \\
\multicolumn{1}{l|}{TA-DualCV$^{-I}$} & 0.397 & 0.337 & 0.522 & 0.399 & 0.471 & 0.516 & 0.380 & 0.416 & 0.409 & 0.427 \\
\multicolumn{1}{l|}{TA-DualCV} & \textbf{0.314}** & 0.222 & \textbf{0.334}** & \textbf{0.303}** & \textbf{0.346}** & 0.295 & \textbf{0.272}** & \textbf{0.267}** & \textbf{0.211}** &  \textbf{0.285}** \\ \hline

\multicolumn{11}{l}{\textbf{Mayo}}  \\ \hline
\multicolumn{1}{l|}{MeanFill}    & 0.710 & 0.680 & 1.998 & \textbf{45.391} & \textbf{0.684} & \textbf{5.217} & \textbf{0.880} & \textbf{1.310} & \textbf{1.662} & 7.109 \\  
\multicolumn{1}{l|}{ECF}     & \textbf{0.640} & \textbf{0.467} & \textbf{0.949} & \textbf{45.391} & \textbf{0.684} & \textbf{5.217} & \textbf{0.880} & \textbf{1.310} & \textbf{1.662} & \textbf{6.942} \\ \cline{2-11}
\multicolumn{1}{l|}{MICE}    & \textbf{0.812} & 0.460 & 0.716 & 24.445 & 0.645 & 0.435 & 0.652 & 0.775 & 0.780 & 3.617\\
\multicolumn{1}{l|}{3D-MICE}  & 0.820 & \textbf{0.452} & \textbf{0.716} & \textbf{24.445} & \textbf{0.546} & \textbf{0.427} & \textbf{0.549} & 0.778 & 0.779 & \textbf{3.592}\\
\multicolumn{1}{l|}{TAME}    & \textbf{0.812} & 1.526 & 1.101 & 31.405 & 0.955 & 1.329 & 0.587 & \textbf{0.730} & \textbf{0.743} & 4.806\\ \hline

\multicolumn{1}{l|}{TA-DualCV$^{-C}$}       & 0.677 & 0.397 & 0.542 & 0.528 & 0.469 & \textbf{0.246}** & 0.490 & 0.692 & 0.702 & 0.527         \\
\multicolumn{1}{l|}{TA-DualCV$^{-I}$} & 0.541 & 0.372 & 0.543 & 0.530 & 0.554 & 0.660 & 0.543 & 0.517 & 0.506 & 0.530 \\
\multicolumn{1}{l|}{TA-DualCV} &  \textbf{0.363}** & \textbf{0.276}** & \textbf{0.321}** & \textbf{0.329}** & \textbf{0.326}** & 0.310 & \textbf{0.316}** & \textbf{0.295}** & \textbf{0.209}** & \textbf{0.305}** \\ \hline

\multicolumn{11}{l}{\textbf{MIMIC-III}}  \\ \hline
\multicolumn{1}{l|}{MeanFill}    & 1.094 & 0.780 & 4.273 & \textbf{0.847} & \textbf{0.946} & \textbf{1.286} & \textbf{1.400} & \textbf{1.345} & \textbf{1.122} & 1.455\\  
\multicolumn{1}{l|}{ECF}     & \textbf{0.375} & \textbf{0.587} & \textbf{3.171} & \textbf{0.847} & \textbf{0.946} & \textbf{1.286} & \textbf{1.400} & \textbf{1.345} & \textbf{1.122} & \textbf{1.231}\\ \cline{2-11}
\multicolumn{1}{l|}{MICE}    & 0.656 & 0.647 & 0.618 & 0.739 & 0.769 & 0.792 & 1.050 & 0.878 & 1.312 & 0.829\\
\multicolumn{1}{l|}{3D-MICE} & 0.607 & 0.642 & 0.684 & \textbf{0.712} & 0.759 & 0.729 & 1.076 & 0.895 & 1.305 & 0.823 \\
\multicolumn{1}{l|}{TAME}    & \textbf{0.287} & \textbf{0.291} & \textbf{0.253} & 0.747 & \textbf{0.246} & \textbf{0.451} & \textbf{0.140}** & \textbf{0.179} & \textbf{0.089}** & \textbf{0.297}\\ \hline

\multicolumn{1}{l|}{TA-DualCV$^{-C}$}     & \textbf{0.228}** & 0.380 & 0.400 & 0.431 & 0.291 & 0.319 & 0.213 & 0.209 & 0.288 & 0.293 \\
\multicolumn{1}{l|}{TA-DualCV$^{-I}$}  & 0.475 & 0.497 & 0.336 & 0.477 & 0.378 & 0.406 & 0.262 & 0.283 & 0.351 & 0.384 \\
\multicolumn{1}{l|}{TA-DualCV} & 0.240 & \textbf{0.278}** & \textbf{0.226}** & \textbf{0.333}** & \textbf{0.223}** & \textbf{0.306}** & \textbf{0.189} & \textbf{0.174}** & \textbf{0.228} & \textbf{0.244}** \\ \hline  
\end{tabular}
\begin{tablenotes}
      \small
      \item - For each block, the best approach is in \textbf{bold}; The best model across ALL is labeled with **.
      \item - DETROIT cannot produce results so we don't include it in the table.
    \end{tablenotes}
\end{threeparttable}
% }
\end{table*}

In this section, we first compare TA-DualCV with baselines in unsupervised learning tasks, imputation, by masking the observations with two commonly-used strategies separately: 

1) Following~\cite{miao2021generative},  we vary random mask rates on 13 general laboratory analyse features from  60\% up to 90\% of observed values using widely-used MIMIC-III so as to evaluate the performance of imputation methods when the missing rate is high and close to the general clinical data. 

2) Following~\cite{luo20183d}, we randomly mask one measurement per feature per visit using 9 sepsis-related features in EHRs from three healthcare systems, CCHS, Mayo, and MIMIC-III, to study the performance of imputation and to btain complete data for supervised learning tasks: septic shock early prediction.

\noindent\textbf{Features} 22 features, including 13 lab analyte and 9 sepsis-related features, are investigated in this study. The 13 lab analyte features are commonly used in existing works \cite{luo20183d, xu2020multi, zhang2020predicting, yin2020identifying}: Chloride, Potassium, Bicarb, Sodium, Hematocrit, Hemoglobin, MCV, Platelet, WBC, RDW, BUN, Creatinine, and Glucose. We use short forms CI, K, HCO3, Na, Hct, Hb, MCV, PLT, WBC, RDW, BUN, Cr, GLC, respectively, to represent these features for simplicity. The 9 sepsis-related features \cite{yang2018time} are from four categories: 1) \textit{Vital Signs}: Temperature, RespiratoryRate, HeartRate; 2) \textit{Respiratory System}: FiO2, PulseOx; 3) \textit{Cardiovascular System}: DiastolicBP, SystolicBP, MAP; 4) \textit{Others}: OxygenFlow. %Missing rates of the features are shown in table \ref{tab:mssingrate}
%\footnote{For simplicity, we use abbreviations: CI, K, HCO3, Na, Hct, Hb, MCV, PLT, WBC, RDW, BUN, Cr, GLC, for 13 lab analyte features; Temp, RespRate, HeartRate, MAP, DBP, SBP, FiO2, PulseOx, OFlow, for 9 sepsis-related features.}

\subsection{Datasets}
\label{sec:imputation-data}
\begin{itemize}
    \item \textbf{MIMIC-III} \cite{johnson2016mimic}\footnote{Dataset is available at http://mimic.physionet.org.} contains admissions of adult patients (i.e., age $>$ 16) who are admitted to intensive care units (ICU) at a tertiary referral hospital between 2001 and 2012, corresponding to 53,423 visits containing $\sim$11 million events. % Moreover, sepsis patients are selected strictly following \cite{zhang2017lstm, yang2018time} which combines International Classification of Diseases, Ninth Revision, and the Third International Consensus Definitions for Sepsis and Septic Shock \cite{singer2016third}. Considering the imbalanced ratio between shock positive and negative visits, we further employ stratified random sampling on both positive and negative visits so that the final numbers of shock positive and negative visits are equal in each dataset. With the selected 9 sepsis-related features, 772 sepsis visits are obtained. The lengths of time series range from 1 to 923. The native missing rate across-visit is $38\%$ on average.     
    \item \textbf{Christiana Care Health System (CCHS)} contains visit records of adult patients (i.e., age $>$ 18) admitted to the Christiana Care hospital from July 2013 to December 2015, corresponding to 210,289 visits containing $\sim$9 million events. %In this work, 1,124 sepsis visits are obtained. The lengths of time series range from 1 to 827. The native missing rate across-visit is $88\%$ on average.
    \item \textbf{Mayo} contains visit records of adult patients (i.e., age $>$ 18) admitted to Mayo Clinic Hospital from July 2013 to December 2015, corresponding to 121,019 visits containing $\sim$52 million events. %In this work, 2,176 sepsis visits are obtained. The lengths of time series range from 1 to 4,863. The native missing rate across-visit is $86\%$ on average.
\end{itemize}

\subsubsection{Data Preprocessing for Imputation with Varied Masking Rates}
With the selected 13 lab analyte features, to provide sufficient data for evaluation with high masking rates and to meet the base implementing requirements of baselines such as those that need a sliding window, we apply exclusion criteria for data sampling in MIMIC-III: i) exclude visits that have less than 11 events; ii) exclude events if they contain more than half missing measurements. Thus, 19,693 visits containing 658,690 events are obtained from MIMIC-III. %The number of events across visits varies from 11 to 1191.

\begin{figure}[t]
\centerline{\includegraphics[scale=0.42]{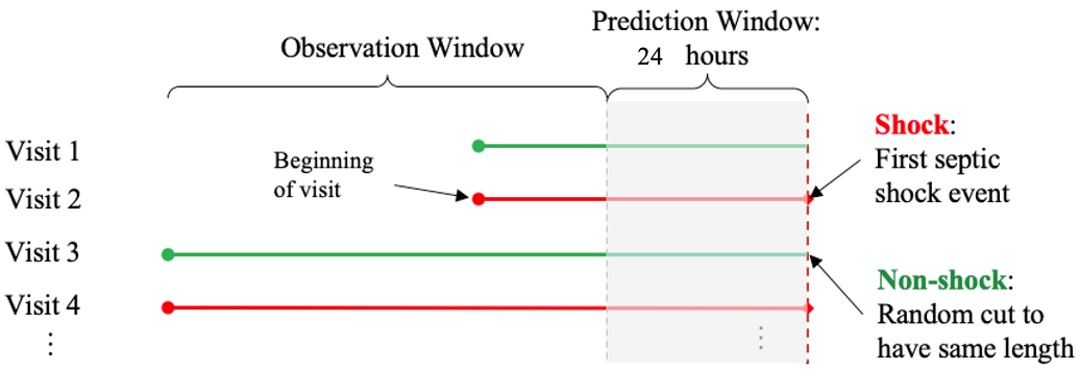}}
\caption{Illustration of septic shock early prediction setting.}
\label{fig:rightalign}
\end{figure}

\subsubsection{Ground Truth Labeling \& Data Preprocessing for Septic Shock}
\vspace{0.05in}
Supervised learning models depend heavily on the accurate label of the training set. However, acquiring the true label (i.e., septic shock and non-shock) can be challenging. Although diagnosis codes, such as International Classification of Diseases, Ninth Revision (ICD-9), are widely used for clinical labeling, solely relying on ICD-9 can be problematic as it has been proven to have limited reliability due to the fact that its coding practice is used mainly for administrative and billing purpose. Indeed, it has been widely argued that ICD-9 cannot be used for establishing reliable gold standards for various clinical conditions \cite{sepsis:ho, zhang2017lstm}. More importantly, ICD-9 cannot tell when septic shock occurs at event level, which is essential for our task. On the basis of the Third International Consensus Definitions for Sepsis and Septic Shock \cite{singer2016third}, our domain experts identified septic shock as any of the following conditions are met: 
\vspace{0.05in}
\begin{itemize}
\item Persistent hypertension as shown through two consecutive readings ($\le$ 30 minutes apart).
    \SubItem{Systolic Blood Pressure (SBP) $<$ 90 mmHg}
    \SubItem{Mean Arterial Pressure (MAP) $<$ 65 mmHg}
    \SubItem{Decrease in SBP $\geq$ 40 mmHg with an 8-hour period}
\item Any vasopressor administration. 
\end{itemize}
\vspace{0.05in}
When combing both ICD-9 and the domain experts' rules, we identify: i) 4,918 visits (2,459 shock positive visits and 2,459 negative visits) containing $\sim$1 million events from MIMIC-III. ii) 144,119 visits (2,963 shock positive visits and 141,156 negative visits) containing 795,314 events from CCHS. iii) 84,897 visits (3,499 shock positive visits and 81,398 negative visits) containing 7,612,360 events from Mayo. We are given all of a patient's EHRs until 24 hours before the septic shock onset (shock group) or end of the sequences (non-shock group), and supervised learning task is to use the complete data to predict whether or not the patient will develop septic shock exactly 24 hours later. To conduct this task, we \emph{right aligned} all the shock sequences by septic shock onset and all non-shock by the end of their sequences and include all the EHRs until 24 hours before the end of sequences (see Fig.~\ref{fig:rightalign}). Our \emph{prediction window} here is 24-hour window before the onset of septic shock or end of sequence. To simulate early prediction in practice, we exclude events in prediction window, accordingly the visits that contain no events in observation window are excluded for both imputation and septic shock prediction tasks. As a result, the final datasets have: i) 772 visits (386 shock positive visits and 386 negative visits) containing 37,246 events from MIMIC-III. Notably, MIMIC-III is ICU-specific and sepsis onset can present itself early, thus the number of remaining visits in MIMIC-III is relatively small after applying a 24-hour cut before the onset. ii) 2,842 visits (1,347 shock positive visits and 1,495 negative visits) containing 376,887 events from CCHS. iii) 6,836 visits (3,457 shock positive visits and 3,379 negative visits) containing 1,547,893 events from Mayo. Note that the labels are not used for imputation tasks, but only for septic shock prediction tasks in Section~\ref{sec:task2}.

% Notably, for nine sepsis-related features imputation, we don't apply strict exclusion criteria as the ones described in former imputation task with 13 lab features, because we would like to investigate imputation performance with native data characteristics, where missing rates can be relatively high and lengths of trajectories can be varied a lot.

\vspace{0.1in}
\subsubsection{Missing Data Analysis}
\vspace{0.05in}
Since different features are measured at different events, plenty of missing entries exist in EHRs. For instance, vital signs are generally measured every 8 hours, while lab values are measured every 24 hours. Hence there may not be available readings for lab results when a new event is created for vital signs. Table~\ref{tab:mssingrate} shows both cross-visit and within-visit missing rates of 22 features selected from the three datasets in our study. We also calculate median and mean value of cross-visit and within-visit missing rates across features. The missingness across visits within one medical system and across different medical systems can diff a lot.

\subsection{Baselines}
\label{sec:imputation-baseline}

We compare TA-DualCV with six imputation approaches for comparison, which are originally designed towards either supervised or unsupervised learning tasks. \footnote{Note that Missing indicator (MI) is not included as a baseline in this task, as it only indicates whether a measurement is missing without imputation.}  \\
\textit{Approaches towards supervised learning tasks:}
\begin{itemize}
    \item \textbf{MeanFill} imputes missing values by mean values cross-visit on each feature.
    \item \textbf{Expert Carry Forward (ECF)} \cite{kim2018temporal} fills the missing values by the last value within a fixed-length window (i.e., 24 hours for lab analytes and 8 hours for vital signs). Then it fills the remaining with the mean value of the corresponding features. ECF is a strong method for both imputation and disease diagnosis in~\cite{kim2018temporal}. 
\end{itemize}
%ECF is a strong method for both clinical data imputation and prediction in existing work \cite{kim2018temporal}. 
\textit{Approaches towards unsupervised learning tasks:}

\textit{Non-NN-Based:}
\begin{itemize}
    \item \textbf{MICE} \cite{buuren2010mice} imputes missing values by capturing multivariate dependencies cross-visit.
    \item \textbf{3D-MICE} \cite{luo20183d} imputes missing values by integrating multivariate dependencies cross-visit and time-interval dependencies within-visit.
    \end{itemize}
    
\textit{NN-Based:}
\begin{itemize}
    \item \textbf{DETROIT} \cite{yan2019detroit} prefills missing values with means and impute missing values based on neural networks by leveraging multivariate and time-aware dependencies from observed values within a 5-length window. 
    \item \textbf{TAME} \cite{yin2020identifying} imputes missing values based on bidirectional RNNs and within-visit multi-modal embedding that takes data including demographics, diagnosis, medication, features, and time-intervals as inputs.     
\end{itemize}
% \textbf{CATSI} \cite{yin2020context} imputes missing values from a fusion layer by integrating multivariate and time-aware dependencies cross-visits. \\

% \textbf{TA-DualCV} is our proposed model for missing values imputation. 
To evaluate the performance of our proposed modules, we implement another two variant versions of TA-DualCV. 
\begin{itemize}
  \item \textbf{TA-DualCV$^{-C}$} removes DualCV but imputes based on within-visit time-aware information.
  \item\textbf{TA-DualCV$^{-I}$} removes within-visit time-aware augmentation that is used to augmenting results from DualCV. 
\end{itemize}

\subsection{Experimental Setup}
To obtain results of baselines, for ECF, we implement the code strictly following the instructions in the authors' paper. For MICE, we use R package \textit{mice} with 10 iterations. For 3D-MICE, DETROIT and TAME, we use the source code provided by the authors with their default parameter settings as suggested in~\cite{miao2021generative}. We implement our proposed work with R and Python. %All experiments are conducted on a machine with two 6-core Xeon E5-2620 CPUs $@2$ GHz and four NVIDIA TITAN Xp / 12GB RAM, running on CentOS 7.8 (64-bit). 
For parameters settings, we use 10 iteration with multiple chain equations. In compromising layer, we use $w_1=w_2=0.5$ for all experiments, to provide a general view of TA-DualCV's performance taking balanced efforts from both cross-visit and within-visit imputation in this study.

\subsubsection{Evaluation Metric}
Following~\cite{luo20183d,yin2020identifying}, normalized root-mean-square error (nRMSE) is used to evaluate the imputation performance of the methods.
We evaluate the imputation performance of all methods on masked values by calculating a popular matrix, normalized root-mean-square error (nRMSE). nRMSE assigns large residuals with a disproportionately large effect, thus more sensitive to outliers \cite{pontius2008components}. 
nRMSE is widely adopted in clinical data imputation tasks that contain different scales of values \cite{luo20183d, yin2020identifying}. As normalization of RMSE has different calculations in different approaches. In this study, we used the common choice of range normalization as in~\cite{luo20183d}.

% The nRMSE of a feature $d\inD$ is calculated as:
% \begin{equation}
%     nRMSE_d = \sqrt{\frac{\sum_{i,t}(\frac{(\hat{Y}_{d,i,t} - Y_{d,i,t})}{max(Y_{d,i}) - min(Y_{d,i})})^2}{\mid T \mid}}
% \end{equation}

% \noindent where $i \in [1, N]$, $t \in \sum_i{X_{i,\cdot,d}^{imp}}$. $N$ is the total number of visits in the data, $T$ is a list of time points with missing values of feature $d$ for patient $i$. $\hat{Y}$ represents the estimated value while $Y$ represents the actual value.

\begin{figure}[t]
\centerline{\includegraphics[scale=0.3]{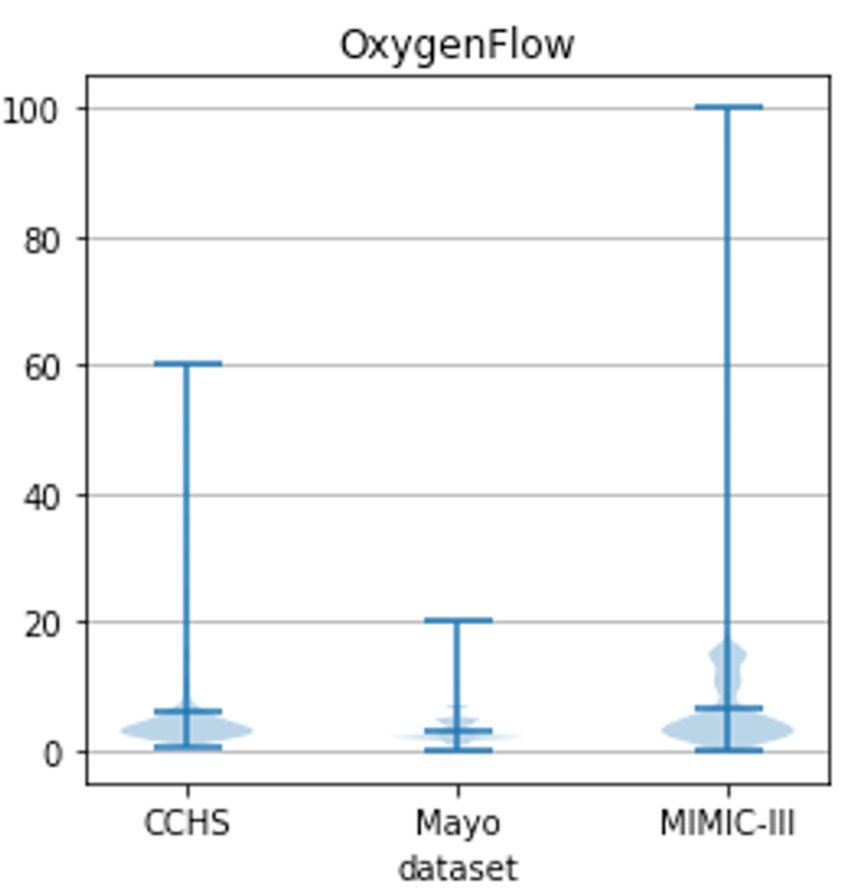}}
\caption{Value distribution of OxygenFlow on CCHS, Mayo, and MIMIC-III. The blue-shaded area represents the density of the data. Blue-vertical lines represent the maximum, mean, and minimum values of the data.}
\label{fig:violindist}
\end{figure}

\subsection{Results}
\label{sec:imputation-result}

Table \ref{tab:nrmse}  compares the various approaches' average nRMSE of 13 general lab analytes features across different masking rates on MIMIC-III. The best approaches for each block is indicated in \textbf{bold}; the best approaches across all blocks is indicated by **. In approaches towards supervised learning tasks, ECF outperforms MeanFill at all masking rates and performance holds steady. For approaches towards unsupervised learning tasks, in non-NN-based approaches, 3D-MICE outperforms MICE at all masking rates. In NN-based approaches, TAME can outperform DETROIT when masking rate is as high as 90\%, while DETROIT outperforms TAME when masking rate is 60\%. The nRMSE of DETROIT is increased faster when masking rate increased. A possible reason is that DETROIT needs to prefill missing values, which can introduce more bias when masking rate is high. 3D-MICE achieves the best performance among four unsupervised learning imputation methods. To our surprise, DETROIT and TAME, especially DETROIT, perform worse than 3D-MICE. There is a possible reason that the datasets in our experiments are highly sparse across visits. DETROIT predicts missing values according to neighboring measurements, so it may be difficult to obtain reliable results when the data is highly sparse and even fails to provide results when the number of events is small (e.g., 1). TAME can take in multi-modal information including demographics, diagnosis, medication, and lab analytes, and it has achieved superior performance in its original settings using multi-modal inputs containing 29 features with 54\% average missing rate, as described in~\cite{yin2020identifying}. However, its performance can suffer when highly sparse 9 features are provided, as information is too limited to construct NN model. Finally, Table \ref{tab:nrmse} shows that TA-DualCV achieves the best performance in terms of average nRMSE across all masking rates on this task. 

%Specifically, TA-DualCV wins with the least nRMSE on all 13 features on 90\% and 80\% mask rates, and 11 features on 70\% mask rate. TA-DualCV is more capable with a high mask rate when it's close to missing rate in general clinical data, with an average of 0.227 nRMSE across features, which is 46\% better than state-of-the-art benchmarks TAME (i.e., 0.424), 90\% better than DETROIT (i.e., 2.380), and 13\% better than 3D-MICE (i.e., 0.260). More importantly, across all features, our proposed TA-DualCV is the least sensitive to the mask rate. Intuitively, TA-DualCV has a smaller range of nRMSE average nRMSE across mask rates. It indicates that TA-DualCV is the most robust to mask rates and features. %By calculating $\hat{V}_{IMDS}$, our proposed evaluation estimator (the lower, the more robust), we have 0.00016 for TA-DualCV, 6.79645 for DETROIT, 0.06978 for TAME, and 0.00019 for 3D-MICE. It indicates that TA-DualCV is the most robust to mask rates and features, which is identical to statistical trends in Table \ref{tab:nrmse}.  
On the CCHS, Mayo, and MIMIC-III datasets, Table \ref{tab:nrmse-sepsis}  provides the overall average nRMSE across nine sepsis-related features for the compared methods. Across all three medical systems, these approaches performances varied considerably. One explanation for this could be that the features are distributed differently in the three medical systems (e.g., OxygenFlow). ECF scores better than MeanFill across the three medical systems. 3D-MICE again performs well among four unsupervised learning imputation methods on CCHS and Mayo, while TAME performs well on MIMIC-III. Note that we were unable to make DETROIT converge to produce results. Moreover, 3D-MICE outperforms ECF across all three medical systems, and all methods perform better on CCHS and MIMIC-III than Mayo. Finally, Table \ref{tab:nrmse-sepsis} shows that TA-DualCV outperforms all baselines as it not only achieves the least overall nRMSE among all three medical systems but its performance is relatively stable. 

In summary, Table \ref{tab:nrmse} and Table \ref{tab:nrmse-sepsis}  show that TA-DualCV is capable of handling a diverse set of feature imputation tasks in both general and ICU-specific real-world medical systems when missing rate is high and achieving state-of-the-art performance.

\section{Septic Shock Early Prediction}
\label{sec:task2}

\begin{table*}[h!]
\centering
\caption{Septic Shock 24-Hour Early Prediction Results (mean$\pm$std) using 9 sepsis-related features on CCHS, Mayo, and MIMIC-III.}
\label{tab:pred}
\resizebox{\textwidth}{!}{%
\begin{threeparttable}
\begin{tabular}{l|lllll|lllll}
\hline
Method & \multicolumn{5}{c|}{Raw}                                                                                                                                       & \multicolumn{5}{c}{Combined with Missing Indicators}                                                                                                \\ \cline{2-11} 
                        & \multicolumn{1}{c}{Accuracy}        & \multicolumn{1}{c}{Precision}          & \multicolumn{1}{c}{Recall}       & \multicolumn{1}{c}{F1}              & \multicolumn{1}{c|}{AUC}            & \multicolumn{1}{c}{Accuracy} & \multicolumn{1}{c}{Precision} & \multicolumn{1}{c}{Recall} & \multicolumn{1}{c}{F1}     & \multicolumn{1}{c}{AUC}    \\ \hline

\multicolumn{11}{l}{\textbf{CCHS}}   \\ \hline                     
MeanFill    & \textbf{.632$\pm$.025}	& \textbf{.637$\pm$.032}	 & \textbf{.611$\pm$.036}  & 	\textbf{.623$\pm$.036}  &	\textbf{.679$\pm$.021}	&	\textbf{.682$\pm$.019}	&  \textbf{.673$\pm$.031}  &	\textbf{.710$\pm$.058}	&  \textbf{.689$\pm$.058}  &	\textbf{.740$\pm$.026}\\  
ECF     &	.631$\pm$.023  &	.636$\pm$.024   & 	.602$\pm$.079   &	.617$\pm$.079	&   .676$\pm$.023	 &	.673$\pm$.020   &	.668$\pm$.018	&   .687$\pm$.044	&   .677$\pm$.044	&  .729$\pm$.023				\\ \cline{2-11}
MICE    &  \textbf{.672$\pm$.020}    &	\textbf{.670$\pm$.026}	&   \textbf{.680$\pm$.048}$^\ddag$	&   \textbf{.673$\pm$.048}   &	\textbf{.729$\pm$.033}	&	\textbf{.708$\pm$.043}	&   \textbf{.713$\pm$.059}	&   .692$\pm$.050	&   \textbf{.703$\pm$.050}	&  \textbf{.770$\pm$.038}$^\ddag$				\\
3D-MICE  &	.666$\pm$.024	&    .673$\pm$.015	&   .643$\pm$.058	&   .656$\pm$.058	&   .726$\pm$.034	&	.704$\pm$.049	&   .707$\pm$.056	&   .692$\pm$.086	&   .697$\pm$.086	& .766$\pm$.054				\\
TAME    &  .482$\pm$.010  & 	.291$\pm$.238   &	.600$\pm$.490	&   .392$\pm$.490   &	.513$\pm$.039   &	.643$\pm$.032   &	.633$\pm$.048	&   \textbf{.714$\pm$.074}$^\ddag$   &	.666$\pm$.074   &	.707$\pm$.016				\\ \hline

TA-DualCV$^{-C}$   &  .679$\pm$.024   &	.688$\pm$.032   &	.652$\pm$.055   &	.668$\pm$.055   &	.726$\pm$.026   &	.714$\pm$.017$^\ddag$   &	.714$\pm$.018$^\ddag$   &	.711$\pm$.027   &	.712$\pm$.027$^\ddag$   &	.769$\pm$.025				              \\
TA-DualCV$^{-I}$  &	\textbf{.690$\pm$.025}**   &  .691$\pm$.024$^\ddag$   &	\textbf{.689$\pm$.051}**   &	\textbf{.689$\pm$.051}**   &	.730$\pm$.036$^\ddag$   &	.701$\pm$.033   &	.701$\pm$.041   &	.701$\pm$.039   &	.700$\pm$.039   &	.765$\pm$.044				\\
TA-DualCV &	.687$\pm$.037$^\ddag$  & \textbf{.700$\pm$.033}**   &	.662$\pm$.034   &	.674$\pm$.034$^\ddag$  &	\textbf{.741$\pm$.020}**   &	\textbf{.721$\pm$.035}**   &	\textbf{.716$\pm$.039}**	&   \textbf{.729$\pm$.044}**   &	\textbf{.721$\pm$.055}**   &	\textbf{.777$\pm$.031}**				\\ \hline

\multicolumn{11}{l}{\textbf{Mayo}}  \\ \hline
MeanFill    &   .695$\pm$.020   &	\textbf{.699$\pm$.033}   &	.690$\pm$.025   &	.693$\pm$.025   &	\textbf{.783$\pm$.016}   &	.722$\pm$.020   &	.711$\pm$.039   &	\textbf{.752$\pm$.055}	&   .729$\pm$.055   &	.805$\pm$.014				\\  
ECF     &   \textbf{.701$\pm$.014}   &	.699$\pm$.023   &	\textbf{.705$\pm$.026}   &	\textbf{.702$\pm$.026}   &	.770$\pm$.014   &	\textbf{.733$\pm$.021}   &	\textbf{.730$\pm$.033}   &	.739$\pm$.029   &	\textbf{.734$\pm$.029}   &	\textbf{.816$\pm$.022}				 \\ \cline{2-11}
MICE    &	\textbf{.711$\pm$.014}  &	\textbf{.715$\pm$.015}  &	.701$\pm$.044  &	\textbf{.707$\pm$.044}  &	\textbf{.786$\pm$.016}  &	.727$\pm$.009  &	.726$\pm$.027  &	.735$\pm$.032  &	.729$\pm$.032  &	.814$\pm$.011				\\
3D-MICE  & 	.697$\pm$.016  &	.706$\pm$.030  &	.673$\pm$.036  &	.689$\pm$.036  &	.778$\pm$.007  &	\textbf{.735$\pm$.009}  &	\textbf{.732$\pm$.023}  &	.743$\pm$.018  &	.737$\pm$.018  &	\textbf{.819$\pm$.009}				\\
TAME    &   .515$\pm$.034   &	.523$\pm$.029   &	.576$\pm$.268   &	.510$\pm$.268   &	.549$\pm$.028   &	.730$\pm$.017   &	.710$\pm$.033   &	\textbf{.780$\pm$.041}**   &	\textbf{.742$\pm$.041}   &	.787$\pm$.010				\\ \hline

TA-DualCV$^{-C}$ &	.713$\pm$.021	&   .717$\pm$.035$^\ddag$   &	.703$\pm$.043   &	.709$\pm$.043   &	.793$\pm$.018	&	.735$\pm$.016   &	.735$\pm$.028   &	.739$\pm$.050   &	.735$\pm$.050   &	.819$\pm$.013				                 \\
TA-DualCV$^{-I}$  &	\textbf{.722$\pm$.014}**	&   .716$\pm$.007   &	\textbf{.733$\pm$.052}**   &	\textbf{.724$\pm$.052}**   &	\textbf{.796$\pm$.013}**   &	.746$\pm$.006$^\ddag$  &	.740$\pm$.028$^\ddag$   &	\textbf{.762$\pm$.047}$^\ddag$    &	.749$\pm$.047$^\ddag$   &	\textbf{.834$\pm$.005}**				\\
TA-DualCV &	.717$\pm$.017$^\ddag$	&   \textbf{.732$\pm$.022}**   &	.724$\pm$.026$^\ddag$   &	.719$\pm$.026$^\ddag$  &	.795$\pm$.018$^\ddag$	&	\textbf{.752$\pm$.009}**   &	\textbf{.760$\pm$.023}**   &	.751$\pm$.043   &	\textbf{.750$\pm$.043}**   &	.830$\pm$.010$^\ddag$				\\ \hline

\multicolumn{11}{l}{\textbf{MIMIC-III}}  \\ \hline
MeanFill    &	\textbf{.816$\pm$.037}   &	.836$\pm$.039   &	.789$\pm$.084   &	.808$\pm$.084   &	.889$\pm$.028	&	\textbf{.848$\pm$.025}   &	.839$\pm$.036   &	\textbf{.863$\pm$.047}   &	\textbf{.850$\pm$.047}   &	.901$\pm$.010				\\  
ECF     &	.834$\pm$.041	&   \textbf{.848$\pm$.052}	&   \textbf{.813$\pm$.061}   &	\textbf{.829$\pm$.061}   &	\textbf{.899$\pm$.023}	&	\textbf{.845$\pm$.034}	&   .845$\pm$.022   &	.846$\pm$.088   &	.842$\pm$.088   &	\textbf{.907$\pm$.018}$^\ddag$				\\ \cline{2-11}
MICE    &	\textbf{.837$\pm$.037}   &	\textbf{.850$\pm$.059}$^\ddag$  &	\textbf{.826$\pm$.046}   &	\textbf{.835$\pm$.046}   &	.887$\pm$.026	&	\textbf{.846$\pm$.020}   &	\textbf{.855$\pm$.028}$^\ddag$   &	.836$\pm$.061   &	.844$\pm$.061   &	.904$\pm$.011				\\
3D-MICE  &	.825$\pm$.015	&   .831$\pm$.053   &	.823$\pm$.045   &	.824$\pm$.045   &	\textbf{.890$\pm$.017}	&	.846$\pm$.022   &	.847$\pm$.024   &	.847$\pm$.050   &	\textbf{.846$\pm$.050}   &	\textbf{.906$\pm$.014}				\\
TAME    & 	.510$\pm$.025   &	.616$\pm$.196   &	.493$\pm$.349   &	.436$\pm$.349   &	.581$\pm$.058	&	.782$\pm$.144   &	.739$\pm$.128   &	\textbf{.957$\pm$.042}   &	.825$\pm$.042   &	.845$\pm$.142				\\ \hline

TA-DualCV$^{-C}$  &   .825$\pm$.021   &   .846$\pm$.036   &	.796$\pm$.036   &	.819$\pm$.036   &	.892$\pm$.021	&	.845$\pm$.032   &	.851$\pm$.034   &	.839$\pm$.072   &	.842$\pm$.072   &	.905$\pm$.015				 \\
TA-DualCV$^{-I}$  &	.838$\pm$.035$^\ddag$   &	.844$\pm$.030   &	.829$\pm$.058$^\ddag$   &	.836$\pm$.058$^\ddag$   &	.900$\pm$.025$^\ddag$	&	.854$\pm$.021$^\ddag$   &   .847$\pm$.037   &	.863$\pm$.038	&   .854$\pm$.038$^\ddag$   &	.903$\pm$.015				\\
TA-DualCV     &   \textbf{.856$\pm$.024}**   &	\textbf{.852$\pm$.028}**   &	\textbf{.864$\pm$.034}**   &	\textbf{.858$\pm$.034}**   &	\textbf{.905$\pm$.018}**   &	\textbf{.863$\pm$.032}**   &	\textbf{.856$\pm$.026}**   &	\textbf{.880$\pm$.045}$^\ddag$   &	\textbf{.864$\pm$.049}**   &	\textbf{.909$\pm$.011}**				\\ \hline  
\end{tabular}
\begin{tablenotes}
      \small
      \item - For each block, the best model is in \textbf{bold}; The best and the second-best models across ALL are labeled with ** and \ddag, respectively.
    \end{tablenotes}
\end{threeparttable}
}
\end{table*}

To examine how our proposed approach performs on a supervised task, we utilize the complete data with nine sepsis-related features generated by implementing different imputation methods, to build an LSTM network to predict an extremely challenging condition in practice, septic shock. We predict septic shock 24 hours before its onset, as suggested by domain experts to enable early treatments that can prevent 80\% of deaths \cite{DBLP:conf/bigdataconf/SohnPC20, henry2015targeted, zhang2019time}.
Furthermore, we combine each imputation method with missing indicator (MI), for prediction and exploit \textit{to what extent the missing pattern information can affect the prediction results}. The missing indicator $\mathbf{MI}_{i,j,k}$ of a missing value $X_{i,j,k}$ is represented as
\begin{center}
    %$MI_{i,j,k}$ = $\begin{cases} 1 &$ if $X_{i,j,k}$ is missing $\\ 0 & otherwise \end{cases}$
    $MI_{i,j,k} = \mathds{1}\{X_{i,j,k}$ is missing$\}$.
\end{center}

\subsection{Experimental Setup}
In this experiment, we use the three datasets that containing events within their observation window as described in Section~\ref{sec:imputation-data} and the same baselines as the ones from Section~\ref{sec:imputation-baseline} except DETROIT, as it does not produce complete data as described in Section~\ref{sec:imputation-result}.
%\subsubsection{Visits Cut and Pad} 
% When conducting the septic shock early prediction, we are given EHRs of a patient's visit until 24 hours before an endpoint to predict whether or not this patient will develop septic shock 24 hours later. For septic shock visits, the endpoint is the onset of septic shock; whereas for non-septic shock visits, the endpoint is a randomly cut point such that the distribution of their length is similar to the shock visits. As shown in Figure~\ref{OWFW}, from the start of a visit until 24 hours before the endpoint is denoted as \emph{observation window}, while the 24-hour leading up to the endpoint is denoted as \emph{prediction window}. 
For parameters settings, in LSTM, we use a general setting with 1 hidden layer with 128 hidden neurons, 0.005 initial learning rate. We adapt Adam optimizer \cite{kingma2014adam} with 64 batches and 25 epochs.

\subsubsection{Evaluation Matrix}
Metrics of accuracy (Acc), recall, precision (Prec), F-score (F1) and area under the ROC curve (AUC) were employed for evaluating our models. Accuracy is the proportion of patients whose labels are correctly identified. Recall indicates what proportion of patients that actually have septic shock can be correctly diagnosed by the model as septic shock. Precision tells what proportion of patients who are diagnosed as septic shock actually have septic shock. F1 is the harmonic mean of precision and recall that sets their trade-off. AUC calculates the tradeoff between recall and specificity. Experiment results are averaged from 5-fold cross-validation.

\subsection{Results}

Table \ref{tab:pred} presents the 24-hours early prediction results of septic shock without and with missing indicators. 
Either without or with missing indicators, there is no clear winner between the two approaches towards supervised learning tasks. Among the three approaches towards unsupervised learning tasks, in general, MICE and 3D-MICE perform better than TAME. Interestingly, TAME performs worse than other baselines without missing indicators, probably because it outputs tensor-shape data and some useful information regarding septic shock prediction is discarded. Across Table \ref{tab:pred},  TA-DualCV is the best approach across all approaches and MI settings. 
%Specifically, the three TA-DualCV-based models without MI generally outperform all baselines in that for each dataset, the highest score across all metrics  generally comes from one of the three TA-DualCV-based models. Among the three TA-DualCV-based models, either TA-DualCV or TA-DualCV$^{-I}$ perform the best on various measures across the datasets. This result indicates that capturing across-visits dependencies within sepsis-related features can effectively help model predictive variables as both TA-DualCV$^{-I}$ and TA-DualCV can capture both multivariate and temporal information across-visits. 
When combined with MI, the performance of each imputation approach is improved compared to the original corresponding approach without MI. This suggests that the missing pattern is indeed an important characteristic of EHRs for prediction. For example, when a patient is in a severe condition, events are likely to be recorded more frequently than when a patient is in a relatively ``healthier" condition. Also, TA-DualCV generally benefits less from MI compared to other methods. For example, on MIMIC-III, the AUC of TA-DualCV only increases by $0.4\%$ after combined with MI, while the AUC of TAME increases by $45\%$. The reason might be TA-DualCV has already captured various dependencies within data, which could contain missing pattern information revealed by MI. Moreover, imputation methods benefit more from MI on CCHS and Mayo, but not much on MIMIC-III probably because the former two contain much more native missing values compared to MIMIC-III.

\section{Discussions \& Conclusion}

In this work, we present \textit{TA-DualCV}, a non-Neural Network-based imputation framework towards both unsupervised and supervised learning tasks in EHRs. TA-DualCV TA-DualCV integrates both cross-visit and within-visit dependencies, by exploiting dependencies among features, time-interval, and time-steps. The robustness of TA-DualCV is evaluated on two tasks: \emph{unsupervised imputation task} and \emph{supervised 24-hour septic shock early prediction task} using EHRs from three different medical systems. In both tasks, TA-DualCV is compared to state-of-the-art baselines. Among the different baselines, 3D-MICE consistently outperformed other baselines on the unsupervised learning tasks, while on the supervised task of early septic shock prediction, there was no clear winner. TA-DualCV, on the other hand, shows its robustness to handle high missing rates of EHRs across medical systems by achieving the best performance on both imputation and septic shock early prediction tasks on all evaluation metrics. 

It is also important to note that we cannot produce results in our experiments with some popular imputation approaches, which are data-driven to a specific type of missingness or data characteristic (e.g., ICU lab analytes data with a small native missing rate). Mayo, for example, can have events ranging from 1 to 5,000 across different visits. Consequently, approaches including Mix-MI \cite{xue2019mixture} and GP-VAE \cite{fortuin2020gpvae} that require tensor-shape inputs cannot be utilized. For approaches relying on sliding windows, including DETROIT\cite{yan2019detroit}, a lower bound on the number of events is needed to compute a sliding window, and a specific density of neighboring measurements must be observed to predict missing values. They are not be able to produce certain results in our experiments. Additionally, our experiments on CCHS and Mayo have fewer training data but higher average native missing rates (up to $73\%$) than those used the existing literature. In practice, as discussed in \cite{xu2020multi}, deep learning relies on high-quality representations of the output of substantive data, whereas their imputation performance dwindles with limited training data.

\section*{Acknowledgements}
This research was supported by the NSF Grants: Generalizing Data-Driven Technologies to Improve Individualized STEM Instruction by Intelligent Tutors (2013502), Integrated Data-driven Technologies for Individualized Instruction in STEM Learning Environments (1726550), CAREER: Improving Adaptive Decision Making in Interactive Learning Environments (1651909), and S.E.P.S.I.S.: Sepsis Early Prediction Support Implementation System (1522107).

We would also like to thank the anonymous reviewers, Xi Yang (North Carolina State University), and Qitong Gao (Duke University) for insightful comments that leads to improved paper presentations.

\bibliographystyle{IEEEtran}
\bibliography{main}

\end{document}